\documentclass[10pt,journal,compsoc]{IEEEtran}

\ifCLASSOPTIONcompsoc
  \usepackage[nocompress]{cite}
\else
  \usepackage{cite}
\fi
\usepackage{amsmath,amssymb,amsfonts}
\usepackage{graphicx}
\usepackage{textcomp}
\usepackage{amssymb}
\usepackage{booktabs}
\usepackage{multirow}
\usepackage{subfigure}
\usepackage{caption}
\usepackage{epstopdf}
\usepackage{subfigure}
\usepackage{changepage}
\usepackage{ulem}
\usepackage{url}
\usepackage{color} 
\usepackage{babel}
\usepackage{booktabs}
\usepackage{threeparttable}
\usepackage[pagewise,switch]{lineno}

\usepackage{algorithm}
\usepackage{algpseudocode}
\usepackage{color}
\definecolor{R_P}{RGB}{0,0,0}
\ifCLASSINFOpdf
\else
\fi

\begin{document}
\title{Unsupervised Local Discrimination for \\Medical Images}
\author{Huai~Chen, Renzhen~Wang, Xiuying~Wang, Jieyu~Li, Qu~Fang, Hui~Li, Jianhao~Bai, Qing~Peng, Deyu~Meng and Lisheng~Wang
\thanks{Huai Chen, Jieyu Li and Lisheng~Wang (corresponding author) are with Institute of Image Processing and Pattern Recognition, Department of Automation, Shanghai Jiao Tong University, Shanghai 200240, P. R. China. E-mail: lswang@sjtu.edu.cn.}
\thanks{Renzhen Wang and Deyu~Meng are with School of Mathematics and Statistics and Ministry of Education Key Lab of Intelligent Networks and Network Security, Xi’an Jiaotong University, Xi’an 710049, P.R.China.}
\thanks{Xiuying Wang  is with the School of Computer Science, The University of Sydney, Sydney, NSW 2006, Australia.}
\thanks{Jianhao Bai and Qing Peng (corresponding author) are with Department of Ophthalmology, Shanghai Tenth People’s Hospital, Tongji University, Shanghai 200240, P.R.China.}
\thanks{Hui Li is with the Cooperative Medianet Innovation Center, Shanghai Jiao Tong University, Shanghai 200240, P. R. China.}
\thanks{Qu Fang is with the Changchun GeneScience Pharmaceutical Co. LTD, P. R. China.}
}
\markboth{Submitted To IEEE Transactions on Pattern Analysis and Machine Intelligence (TPAMI)}
{Huai Chen \MakeLowercase{\textit{et al.}}: Unsupervised Local Discrimination for Medical Images}

\IEEEtitleabstractindextext{%
\begin{abstract}
Contrastive learning, which aims to capture general representation from unlabeled images to initialize the medical analysis models, has been proven effective in alleviating the high demand for expensive annotations. Current methods mainly focus on instance-wise comparisons to learn the global discriminative features, however, pretermitting the local details to distinguish tiny anatomical structures, lesions, and tissues. To address this challenge, in this paper, we propose a general unsupervised representation learning framework, named local discrimination (LD), to learn local discriminative features for medical images by closely embedding semantically similar pixels and identifying regions of similar structures across different images. Specifically, this model is equipped with an embedding module for pixel-wise embedding and a clustering module for generating segmentation. And these two modules are unified through optimizing our novel region discrimination loss function in a mutually beneficial mechanism, which enables our model to reflect structure information as well as measure pixel-wise and region-wise similarity. Furthermore, based on LD, we propose a center-sensitive one-shot landmark localization algorithm and a shape-guided cross-modality segmentation model to foster the generalizability of our model. When transferred to downstream tasks, the learned representation by our method shows a better generalization, outperforming representation from 18 state-of-the-art (SOTA) methods and winning 9 out of all 12 downstream tasks. Especially for the challenging lesion segmentation tasks, the proposed method achieves significantly better performances. The source codes are publicly available at https://github.com/HuaiChen-1994/LDLearning.

\end{abstract}

}


\maketitle
\IEEEdisplaynontitleabstractindextext
\IEEEpeerreviewmaketitle
\IEEEraisesectionheading{\section{Introduction}\label{sec:introduction}}
\IEEEPARstart{A}rtificial intelligence-based computer-aided diagnosis has recently become the mainstay of various fields of precision medicine \cite{ahmed2020artificial}, such as radiotherapy \cite{meyer2018survey} and disease screening \cite{ardila2019end}. Intelligent medical image analysis plays a crucial role in the more efficient and effective diagnosis and is the basis for computer-aided systems \cite{litjens2017survey}. However, the heavy reliance on vast amounts of expensive annotations for establishing a generalized and robust medical image analysis model hinders its broad applications. To address this issue and alleviate the high demand for annotations, contrastive learning, which is the mainstream of self-supervised representation learning \cite{chen2019self,zhu2020rubik}, has been a hot research topic and achieved huge breakthroughs \cite{li2020self,li2021rotation}.

In contrastive learning methods, the contrastive rule for constructing negative and positive pairs is of vital importance. Recently, ingenious rules have been proposed for various medical modalities images, including color fundus images \cite{li2020self,li2021rotation}, chest X-ray images \cite{han2021pneumonia,zhang2020contrastive}, and MRI \cite{chaitanya2020contrastive}. For example, based on the similarity between fundus fluorescein angiography (FFA) images and color fundus images, Li et al. \cite{li2020self} treated the color fundus image and the FFA image of the same patient as a positive pair and otherwise as negative pairs, by which modality-invariant and patient-similarity features were captured to enhance retinal disease diagnosis. Based on the hypothesis that aligned MRI images share similar contents in the same anatomical region, Krishna et al. \cite{chaitanya2020contrastive} managed volumes of similar positions as positive pairs to learn features to improve segmentation tasks. These methods have thoroughly investigated the characteristics of medical images and the demands of tasks, and the proposed models are proven effective for solving the targeting tasks.

Most of the recent work focuses on constructing positive/negative pairs from a global perspective to learn discriminative features \cite{li2021rotation,han2021pneumonia,sowrirajan2021moco}; however, the local discrimination, which is essential for depicting lesions, minor tissues, and tiny structures, has not been sufficiently investigated and exploited. Aiming to learn local features from aligned corresponding positions in the registered images, \cite{chaitanya2020contrastive} relies on image registration and is ineffective for images with severe anatomical deformation. Furthermore, most current techniques are specifically designed for specific imaging modalities or tasks. Correspondingly, particular training data is needed for the modality-specific or task-driven models, which in turn hinders the current contrastive learning models from broad clinical applications. For example, corresponding color fundus images and FFA images are needed in \cite{li2020self}, and image-text pairs are needed in \cite{zhang2020contrastive}.

\begin{figure}[]
	\includegraphics[width=0.97\columnwidth]{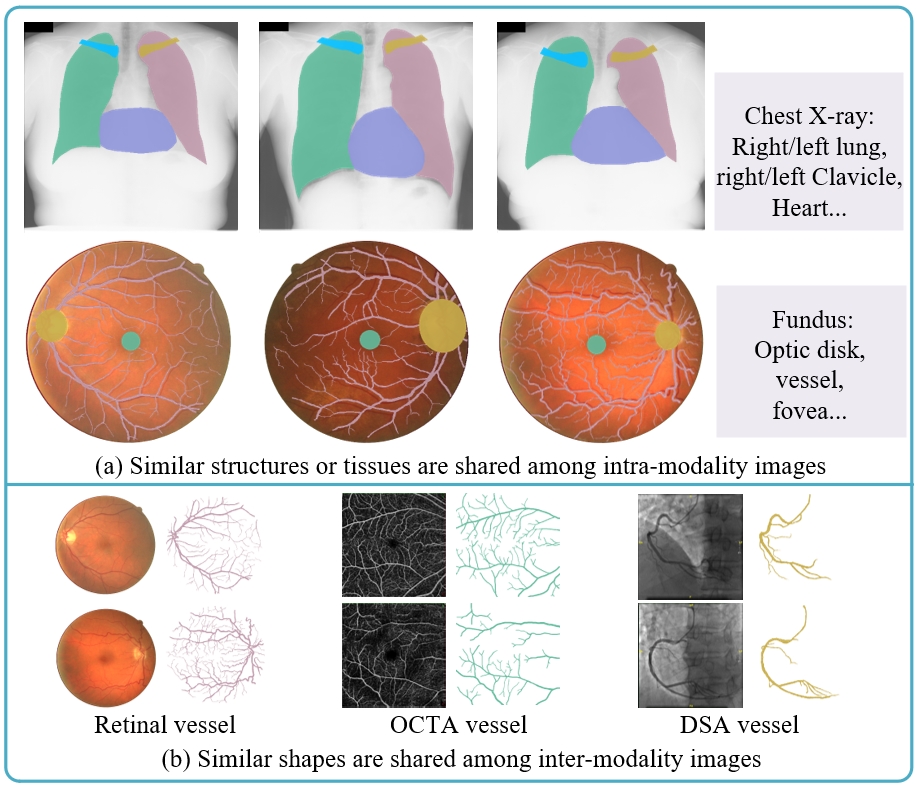}
	\caption{An illustration of the intra-modality structure similarity and the inter-modality shape similarity.} 
	\label{fig:similarity}
\end{figure}
To address the challenge of learning local discriminative representation for medical image analysis, alleviate the high demand for expensive annotations, and ensure that the representation learning framework is generalized for most medical modalities and tasks, we design a universal contrastive rule and propose a general local discrimination learning framework. Concretely, as shown in Fig.~\ref{fig:similarity}.(a), almost all medical images possess the property of intra-modality structure similarity due to the consistency of anatomical structures and the stable focusing region of the imaging modality, which means that images of the same medical modality share structure regions with similar patterns \cite{tortora2018principles}. For instance, each color fundus image has the optic disk (OD), vessels, and fovea. Moreover, the pixels within each structure region are semantically consistent. Therefore, for medical images of the same modality, clustering semantically similar pixels in an image to form structure regions and recognizing regions with similar patterns in different images can be a universal contrastive rule. Based on this rule, we propose region discrimination, which attempts to closely embed semantically consistent pixels and distinguish regions of different semantic categories. By optimizing region discrimination model, local discriminative features that are capable of measuring both pixel-wise and region-wise similarity can be captured.

Furthermore, based on the ability of region discrimination to cluster semantically similar pixels, we make a new perspective to combine region discrimination with the cross-modality shape prior to realize segmentation. As shown in Fig.~\ref{fig:similarity}.(b), although there are obvious gaps in the content and visual characteristics of color fundus images, optical coherence tomography angiography (OCTA) images, and digital subtraction angiography (DSA) images, their vessels have similar shapes. This property is named inter-modality shape similarity and is widely shared among anatomical structures, especially vessels \cite{tortora2018principles}. Based on this property, we can re-utilize the existing manual segmentation masks of similar structures as the reference for the clustering of region discrimination, ensuring that the segmentation region maintains the interior semantic consistency and possess the expected shape, and thereby obtain the segmentation of the target structure.

Finally, to take full advantage of the learned representation that is capable of measuring regional similarity, we propose a one-shot center-sensitive landmark localization model. It introduces the center-sensitive property into the discriminative representation, by which landmarks can be accurately localized with the guidance of one labeled sample. Since the learned features are augmentation-invariant and orientation-invariant, this model is generalized and robust for multiple medical tasks. 

In summary, our major contributions are three folds:

\begin{itemize}
\item[$\bullet$] A universal representation learning framework is proposed to generate representation with the ability to measure pixel-wise and region-wise similarity. The learned representation significantly enhances the medical image models focusing on analyzing tiny structures/tissues and alleviates their demand for expensive annotations. More specifically, our model improves the mean Dice Similarity Coefficient (DSC) by 5.79\% for 8 color fundus tasks and 3.68\% for 4 chest X-ray tasks, respectively.

\item[$\bullet$] A new schema to transfer shape knowledge across medical modalities is presented, which can be fused with our representation learning model to realize shape-guided segmentation. This model demonstrates the feasibility of identifying anatomical structures by harnessing the shape knowledge from similar structures and the interior semantic consistency, and opens up a new mechanism to effectively re-utilize the existing annotations.

\item[$\bullet$] Based on the learned representation with the ability to measure regional similarity, a center-sensitive one-shot landmark localization method is proposed, which makes the landmark localization more economic and convenient. It is one of the pioneering self-supervised one-shot landmark localization frameworks and is generalized for various medical tasks. 

\end{itemize} 

This paper is an extended version of our conference paper \cite{chen2021unsupervised}. The substantial extensions are as follows: 1) This study has innovated the optimization model to enhance the generalization and robustness of the representation learning framework, including proposing region discrimination loss function to capture the comprehensive structure information and alleviate the disturbance of noisy data, and presenting an evolutionary optimization strategy to enhance the stability of simultaneously training the clustering and the embedding. 2) To address the frangibility posed by the absence of definite annotations in the pilot shape-guided segmentation model\cite{chen2021unsupervised}, we have proposed an enhanced framework, including re-clustering refinement to adapt the segmentation model to an individual test instance, and uncertainty estimation refinement to improve model's robustness. 3) We have extended our model to a new clinical application task of landmark localization, and proposed center-sensitive one-shot landmark localization model, which can accurately identify targets with only one labeled sample. 4) Compared with the preliminary version, this version includes more necessary clarifications, discussions, and evaluations on the capabilities of the proposed model.

The following paper is organized as follows: Sec.~\ref{section:related-work} overviews the related work. The preliminary knowledge is elucidated in Sec.~\ref{section:preliminaries}. Sec.~\ref{section:local discrimination learning} presents the details of our representation learning model. Sec.~\ref{section: Shape-guided Segmentation and One-shot Localization} deeply investigates the clinical application of our method for segmentation and localization. In Sec.~\ref{section: experiments}, extensive experiments are conducted to analyze our method quantitatively and qualitatively. Finally, conclusions are summarized in Sec.~\ref{section:conclusion}.

\section{Related Work}
\label{section:related-work}
\subsection{Contrastive Learning}
\textcolor{R_P}{Contrastive learning is a critical category of self-supervised learning. Its core idea is to construct positive/negative pairs based on some similarity measuring rules, and learn a semantic embedding space to map positive pairs closely and negative pairs separately \cite{he2020momentum,chen2020simple,wang2021dense}. Through minimizing contrastive loss, such as InfoNCE \cite{van2018representation}, the models are encouraged to learn discriminative features that can be transferred to enhance downstream tasks. To construct positive and negative pairs, the early work \cite{hadsell2006dimensionality} was based on datasets with prior knowledge, or manual annotations, and treated images with neighborhood relationships as positive pairs and otherwise as negative pairs. More recently, contrastive learning methods based on unlabeled images have been the mainstay, and can be divided into instance-wise contrastive learning \cite{he2020momentum,chen2020simple,wang2021dense} and cluster-wise contrastive learning \cite{caron2020unsupervised,li2020prototypical,li2021contrastive}.}

\textbf{Instance-wise contrastive learning} can be regarded as the extension of exemplar-CNN \cite{dosovitskiy2015discriminative}, in which, the images of the same instance but with varied augmentations are treated as the same category and the discriminative features are learned through the classification procedure. Although this framework is capable of capturing semantic features, it is impractical when training with large-scale datasets due to the high computational demand of adopting a parametric paradigm. To address this issue, more recent instance discrimination methods \cite{wu2018unsupervised,ye2019unsupervised,wang2021dense,chen2020simple} adopt a non-parametric classifier and a noise-contrastive estimation to recognize instances with a fixed computational cost. MoCo \cite{he2020momentum} and MoCov2 \cite{chen2020improved} are two typical instance-wise discriminative methods, in which, an on-the-fly dictionary composed of an embedding encoder and a momentum encoder is built to match queries and keys. They have achieved milestones in closing the performance gap between unsupervised and supervised representation learning. 

In our framework, the preparatory stage of patch discrimination is mostly related to instance-wise contrastive learning. \textcolor{R_P}{However, to satisfy the demand of medical image processing for describing local details, patch discrimination maps an image to dense spatial feature vectors instead of the commonly utilized global feature vector. Each of these feature vectors represents a patch and reflects its comprehensive semantic information by integrating the patch’s inner content and nearby context.} In patch discrimination, a patch instance is treated as an individual category, and the feature measuring patch-wise similarity can be obtained by forcing the model to recognize different views of the same patch instance. Furthermore, hypersphere mixup is presented for patch discrimination to improve the smoothness of embedding space and enhance the generalization of the representation.

\textcolor{R_P}{\textbf{Cluster-wise contrastive learning} \cite{li2021contrastive,caron2020unsupervised,li2020prototypical} attempts to harness clustering and contrastive learning to investigate the inter-image similarity and treat two similar images as a positive pair. For example, Sharma et al. \cite{sharma2020clustering} treated the clustered results as the labels when constructing positive/negative pairs, and successfully learned effective face representation. To enhance the ability on depicting the semantic structure, Li et al. \cite{li2020prototypical} assigned image instances to several prototypes (cluster centroids) of different granularity, and then the contrastive loss was constructed to enforce an image instance and its assigned prototype to have similar embeddings.}

\textcolor{R_P}{Our region discrimination also aims to introduce clustering to enhance contrastive learning. However, rather than clustering images into several cluster assignments in a global perspective, our method focuses on clustering semantically consistent regions based on the intra-modality structure similarity, which contributes to much finer features with the ability to measure both pixel-wise and region-wise similarities. Furthermore, due to the mutually beneficial training mechanism, the clustered results and the embeddings are directly generated by our model without extra clustering stages as K-means in \cite{li2020prototypical} and FINCH in \cite{sharma2020clustering}, which is efficient and economic.}

\subsection{Contrastive Learning in Medical Images} 
\textcolor{R_P}{Contrastive learning has been a hot topic in medical image analysis \cite{chaitanya2020contrastive,han2021pneumonia,zhang2020contrastive}. According to the design mechanisms, recent work can be mainly classified into specific-properties-based methods \cite{li2020self,liu2021simtriplet,yang2021self} and application-driven methods \cite{han2021pneumonia,li2021rotation,zhao2021unsupervised}. \textbf{Specific-properties-based methods} aim to fuse the characteristics specific to the input modality into contrastive learning. For instance, considering the domain-specific knowledge that different dyes can enhance different types of tissue components, Yang et al. \cite{yang2021self} separated the stain results of hematoxylin and eosin, and fused them into the design of contrastive learning. \textbf{Application-driven methods} take account of the properties of the processing tasks when designing models. For instance, based on the observation that the structure information in retinal images is sensitive to the orientations and is beneficial for the recognition of retinal diseases, Li et al. \cite{li2021rotation} fused the rotation prediction task with contrastive learning to improve the classification of pathologic myopia.}

Most of the recent medical contrastive learning methods focus on using instance discrimination to learn global features, while neglecting local details \cite{li2021rotation,han2021pneumonia,zhang2020contrastive,sowrirajan2021moco}. And further, the special requirements for the training data hinder the current contrastive learning from broad medical applications \cite{li2020self,zhang2020contrastive,han2021pneumonia}. Comparatively, our method learns local discriminative representation with the ability to measure pixel-wise and region-wise similarity. And by incorporating the common property of intra-modality structure similarity, our model is capable of processing and analyzing images for multiple medical modalities and downstream tasks.

\subsection{Cross-domain Knowledge Transfer} 
The purpose of cross-domain knowledge transfer is to improve the performance of tasks in the target domain by utilizing the knowledge from related domains. According to differences in the working mechanisms, it can be divided into domain adaptation and synthetic segmentation. \textbf{Domain adaptation} aims to adjust the trained model in the source domain to be generalized in the target domain \cite{wilson2020survey,ahn2020unsupervised}, which effectively alleviates the degeneration of the model caused by the changing imaging conditions.

\textbf{Synthetic segmentation} is a unique cross-domain knowledge transfer framework for medical images to share knowledge across modalities \cite{yang2020unsupervised,8494797,zhang2018translating,zhou2021anatomy}. It is based on the hypothesis that images of medical modalities focusing on the same body part, such as the head CT and the head MRI, share similar organs and tissues. Based on this, synthetic segmentation usually utilizes image translation to get synthetic target modality images from source labeled images. After that, the synthetic images are accompanied by original real labels to train a segmentation network and this model can be directly applied to the target modality. Synthetic segmentation has been widely investigated in cross-modality pairs, such as CT-MR \cite{yang2020unsupervised,8494797}, and CBCT-MR \cite{zhou2021anatomy}. Recently, CycleGAN-based synthetic segmentation, which is an unsupervised framework based on unpaired images, has obtained more attention than supervised ones \cite{yang2020unsupervised,8494797,zhang2018translating,chen2020anatomy}.

Synthetic segmentation is a cost-effective method to transfer knowledge across modalities. It demonstrates that the deep network owns a similar ability as humans to memory and transfer knowledge from familiar medical scenarios to unknown ones \cite{akagunduz2019defining}. However, it can only share knowledge among modalities with almost the same content, i.e., modalities of the same body part. For radiologists, the ability to transfer knowledge is more powerful, by which, the knowledge can still be shared across modalities with big content differences. For example, the knowledge of vessel shape in OCTA or DSA is enough for a radiologist to recognize vessels in color fundus images. The inherent mechanism is that people can recognize a target based on shape knowledge and its interior pattern similarity. Therefore, we attempt to demonstrate the feasibility of transferring shape knowledge among modalities with big content gaps to realize segmentation, and successfully propose shape-guided cross-modality segmentation. 

\subsection{\textcolor{R_P}{Self-supervised One-shot Landmark Localization}}
\textcolor{R_P}{Anatomical landmark localization is essential for medical therapy planning and intervention, and has been a hot research topic \cite{xu2019efficient,zhou2021review}. The supervised localization models often heavily rely on expensive annotations on the images. To alleviate the burden of manual annotating, self-supervised one-shot landmark localization frameworks \cite{lei2021contrastive,yao2021one} have been designed, which train models in easily accessible unlabeled images and only require one labeled sample as supporting information. For example, RPR-Loc \cite{lei2021contrastive} was based on the idea that tissues and organs from different human bodies own similar relative positions and contexts. And thus, this method predicted the relative positions between two patches to locate the target organ. Cascade Comparing to Detect (CC2D) \cite{yao2021one} was a coarse-to-fine localization framework, where coarse-grained corresponding areas were firstly identified and then the finer-grained areas were matched in the selected coarse area. For CC2D, the target point is randomly selected in the original image, and the training target is to detect the best matching point in the cropped image.}

\textcolor{R_P}{Our center-sensitive one-shot landmark localization method is based on our patch discrimination, and utilizes the learned patch discriminative features to search for the best matching patch. To ensure that the detected results are precisely aligned with target landmarks, we further propose the center-sensitive averaging pooling, making the representation of a patch pay more attention to the semantic information in the patch's central regions. Compared with recent self-supervised methods, where CC2D requires most pixels in the training images have structural information and RPR-Loc needs the structures in training images should have stable relative positions, our method is more generalized for fitting different medical modalities since the trained patch-wise embedding is augmentation-invariant and orientation-invariant.}

\section{Preliminaries}
\label{section:preliminaries}

In this section, we firstly articulate the basic mechanism of contrastive learning in Sec.~\ref{subsection:Base_Contrastive_Formulation} and elucidate the concept from instance-wise discriminative features to local discriminative features in Sec.~\ref{subsection:Instance_discrimination_to_Region_discrimination}. 

\subsection{Base Contrastive Formulation}
\label{subsection:Base_Contrastive_Formulation}
Traditional contrastive learning commonly formulates an embedding function that projects images into a $K$-d embedding space to make positive pairs closely clustered and negative pairs dispersedly distributed. We denote $v_i\in\mathbb{R}^K$ as the embedded vector of the $i$-th sample, $v^+\in{V_i^+}$ and $v^-\in{V_i^-}$ as its positive and negative samples, where $V_i^+$ and $V_i^-$ are respectively the sets of positive and negative embedded vectors for $v_i$. The contrastive loss is defined as:
\begin{equation}
	\mathcal{L}_i=\frac{1}{|V_i^+|}\sum_{v^+\in{V_i^+}}-\text{log}\frac{e^{\text{sim}(v^+,v_i)/\tau}}{e^{\text{sim}(v^+,v_i)/\tau}+\sum \limits_{v^-\in{V_i^-}}e^{\text{sim}(v^-,v_i)/\tau}},
\end{equation}where $\text{sim}(a,b)$ means the similarity between two vectors and is usually set as the cosine similarity, i.e., $\text{sim}(a,b)=a^\top b/\left\|a\right\|_2 \left\|b\right\|_2$. And $\tau$ is a temperature parameter to control the concentration level of the distribution \cite{hinton2015distilling}.

\begin{figure}[t]
	\centering	
	\includegraphics[width=0.97\columnwidth]{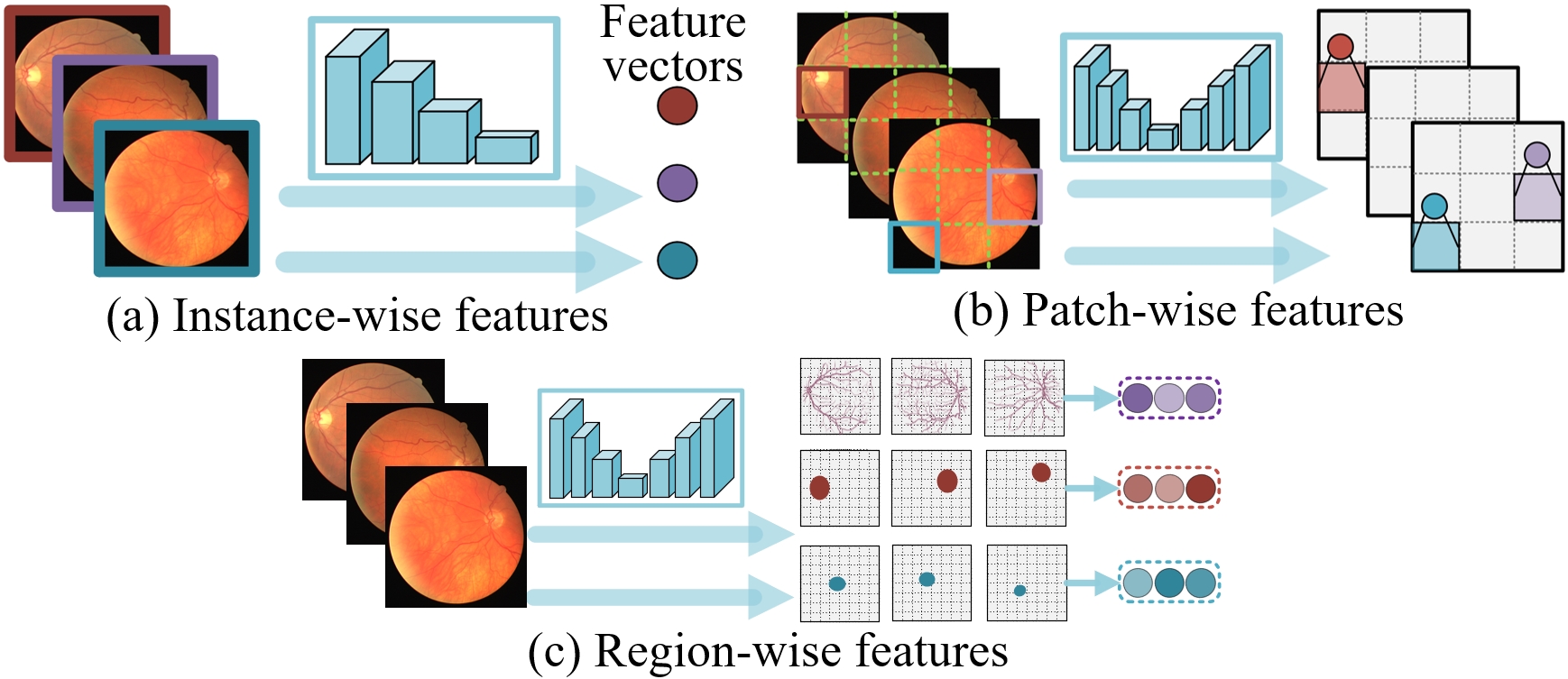}
	\caption{An illustration of instance-wise, patch-wise, and region-wise discriminative features.} 
	\label{fig:instance-patch-region-discrimination}
\end{figure}

\subsection{From Instance-wise Discriminative Features to Local Discriminative Features}
\label{subsection:Instance_discrimination_to_Region_discrimination}

\noindent\textbf{Instance-wise discriminative features:} The common rule of constructing positive and negative pairs is to create multiple views of images with varied augmentations, and a positive/negative pair refers to two views from the same/different images \cite{wu2018unsupervised,he2020momentum,chen2020simple}. Following this rule, instance-wise comparisons are conducted to learn the global representation that can distinguish different image instances. The flowchart is shown in Fig.~\ref{fig:instance-patch-region-discrimination}.(a), where image sample $x_n$ is fed into an encoder $f(\cdot)$ with a global average pooling layer, by which the discriminative information for the whole image can be captured, i.e., $v_{n}=f(x_n)$.

\noindent\textbf{Local discriminative features:} Instance-wise discriminative features are highly concentrated for semantic context; however, medical image tasks, especially segmentation of lesions, tissues, and tiny structures, need finer features describing local details. To address this issue, we attempt to learn local discriminative features, which evolve from patch-wise discriminative features to region/pixel-wise discriminative features. To learn patch-wise discriminative features, dense spatial features are generated to represent patch instances, and each patch is treated as an individual class (shown in Fig.~\ref{fig:instance-patch-region-discrimination}.(b)). As shown in Fig.~\ref{fig:instance-patch-region-discrimination}.(c), region-wise discriminative features are captured by identifying the structure region composed of semantically consistent pixels in an image and clustering regions of the same semantic class among different images. \textcolor{R_P}{Compared with patch-wise discriminative features that can measure patch-wise similarity, region-wise discriminative features, which are capable of reflecting comprehensive structure information as well as measuring pixel-wise and region-wise similarity, better satisfy the demand of medical image analysis for distinguishing tiny structures/tissues.}

\begin{figure}[b]
	\includegraphics[width=0.97\columnwidth]{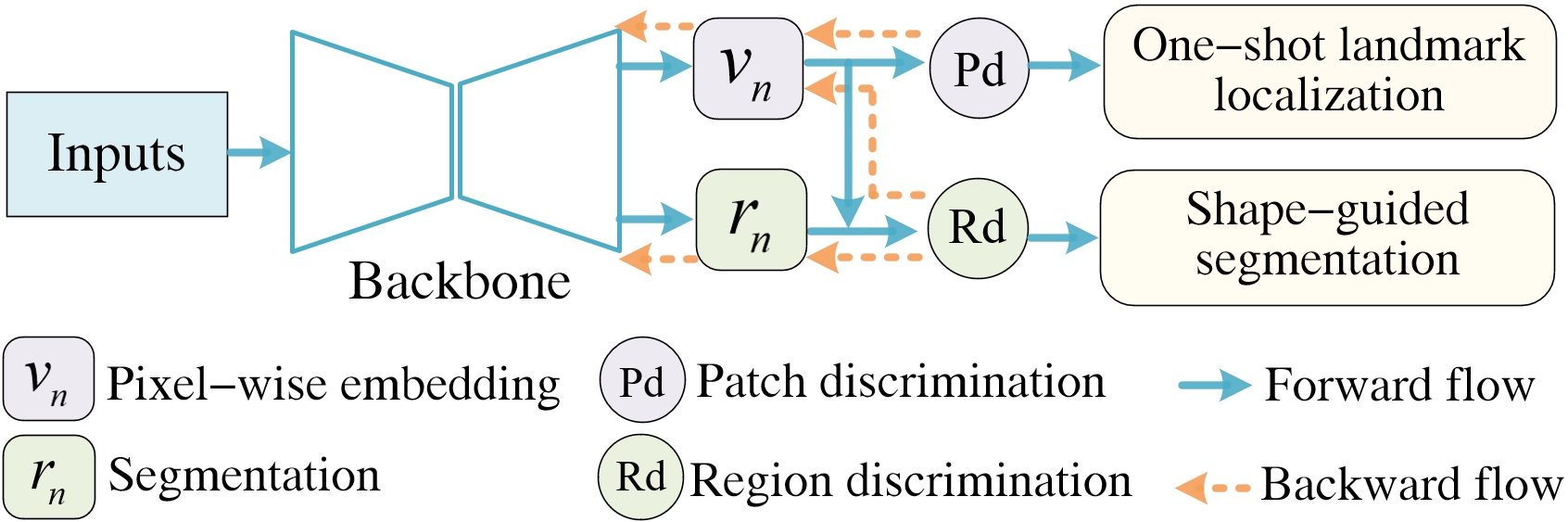}
	\caption{\textcolor{R_P}{The pipeline of the overall schematic framework.}}
	\label{fig:pipeline}
\end{figure}

\noindent \textcolor{R_P}{\textbf{The pipeline of the overall framework:} As shown in Fig.~\ref{fig:pipeline}, our proposed framework is composed of three major parts, including (1) a backbone to generate embeddings and segmentation masks, (2) a systematic local discrimination model, containing patch discrimination and region discrimination, to learn local discriminative features, and (3) the clinical applications of one-shot landmark localization and the shape-guided segmentation.}

\section{Local Discrimination Learning}
\label{section:local discrimination learning}

\begin{figure}[]
	\includegraphics[width=\columnwidth]{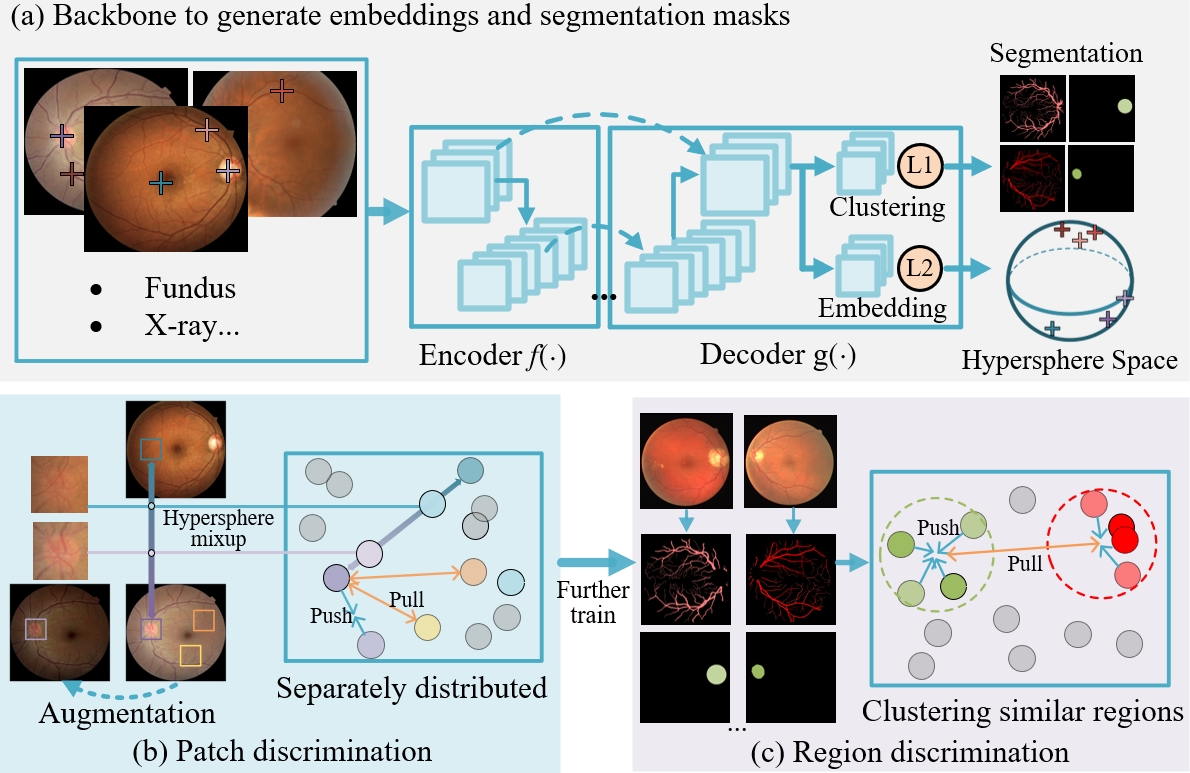}
	\caption{An illustration of local discrimination learning.} 
	\label{fig:framework}
\end{figure}

The illustration of our local discriminative representation learning framework is shown in Fig.~\ref{fig:framework}. It is designed as an evolutionary framework, which firstly utilizes patch discrimination to warm up the embedding module with the initial patch-wise discriminative ability and then further optimizes the total model by region discrimination to learn region-wise and pixel-wise discriminative features. In Sec.~\ref{subsection:backbone}, we present the definition and notation of the backbone. Then, patch discrimination and region discrimination are respectively articulated in Sec.~\ref{subsection:patch-discrimination} and Sec.~\ref{subsection:region-discrimination}. Finally, hypersphere mixup to enhance the generalization of the model is elucidated in Sec.~\ref{subsection:Hypersphere-Mixup}.

\subsection{The Definition and Notation of the Backbone}
\label{subsection:backbone}
\textcolor{R_P}{The backbone aims to generate pixel-wise embedding and segmentation masks, and therefore should support point-to-point mapping. Its design is based on U-Net \cite{ronneberger2015u} that is one of the most commonly utilized segmentation models with the ability to fuse low-level and high-level features.} Specifically, as shown in Fig.~\ref{fig:framework}.(a), the backbone is composed of an encoder $f(\cdot)$ and a decoder $g(\cdot)$. And the final block is composed of a clustering module and an embedding module to realize segmenting and embedding.

We denote the unlabeled image set as $X=\{x_1, x_2,\cdots,x_N\}$ and the image instance as $x_n\in{\mathbb{R}^{H\times W}}$, where $H$ and $W$ are respectively the height and width of this image. After feeding $x_n$ into this model, we can get the segmentation mask $r_n \in \mathbb R^{H\times W\times C}$ and the pixel-wise embedding $v_n \in \mathbb R^{H\times W\times K}$, i.e.:
\begin{equation}
	\label{equation:f_g_2_v_r}
	\begin{aligned}
		v_n&,r_n=g(f(x_n)).
	\end{aligned}
\end{equation}$v_n(h,w)\in{\mathbb{R}^{K}}$ is the embedded vector for pixel $x_n(h,w)$, and $r_{nc}(h,w) (0\leq c<C)$ is the probability of classifying this pixel into $c$-th semantic class. $v_n(h,w)$ is further processed by $l_2$-normalization to project the embedded vector  into a unit space, i.e., $\left\|v_n(h,w)\right\|_2=1$. Meanwhile, $r_n(h,w)$ is processed by $l_1$-normalization to make the sum of classification probabilities of a single pixel to be 1, i.e., $\left\|r_n(h,w)\right\|_1=1$.

\subsection{Patch Discrimination}
\label{subsection:patch-discrimination}
Patch discrimination is the first step to empower our model with the initial discriminative ability. Its core idea is to treat each patch instance as an individual class and force the model to cluster different views of the same patch together. Such a patch-wise contrastive learning scheme enables the model to measure the semantic similarity between patch instances and the resulting model can capture patch-wise discriminative representation.

As shown in Fig.~\ref{fig:framework}.(b), both image sample $x_n$ and its augmented sample $\hat{x}_n$ are fed into the model to get the pixel-wise embedding $v_n$ and $\hat{v}_n$. 
\textcolor{R_P}{Then, the embedded vector for a patch can be denoted as the sum of pixel-wise embedded vectors within the patch.} To simplify the process of constructing pairs and improve the practicability, we evenly divide an image sample into $H_p\times{W_p}$ patches. After that, the embedded feature of the $i$-th patch $p_{ni}$ $(0\leq i<H_p{W_p})$ is denoted as $s_{ni}$ and can be formulated as:
\begin{equation}
	\label{equation:patch_embedding}
	\begin{aligned}
	&\ \ \ \ \  s_{ni}=\frac{\sum_{h\in[h_b,h_e),w\in[w_b,w_e)}v_n(h,w)}{\left\|\sum_{h\in[h_b,h_e),w\in[w_b,w_e)}v_n(h,w)\right\|_2},
	\end{aligned}
\end{equation}where $(h_b,w_b)$ and $(h_e,w_e)$ are respectively the top-left point and the bottom-right point of this patch, and their detailed definitions are as follows:
\begin{equation}
	\label{equation:top_left_bottom_right}
	\begin{aligned}
		&h_b=H/H_p\times \lfloor i/W_p \rfloor ,\qquad \qquad  \ \ h_e=H/H_p+h_b,\\
		&w_b=W/W_p\times (i-W_p\lfloor i/W_p \rfloor) ,\ w_e=W/W_p+w_b,
	\end{aligned}
\end{equation}where $\lfloor \cdot \rfloor$ means rounding numbers down to the nearest integer.

Following the similar analysis as above, we can capture the embedded vector for the corresponding augmented patch in $\hat{x}_n$ and denote it 
as $\hat{s}_{ni}$. Therefore, $S_{ni}^+=\{s_{ni}\}$ is the positive embedding set for $\hat{s}_{ni}$, and the negative embedding set can be denoted as $S_{ni}^-=\{s_{n'i'}|n'\neq n\ or\ i'\neq i\ \}$. Then, the probability of recognizing $\hat{s}_{ni}$ as $s_{ni}$ can be defined as:
\begin{equation}
	\begin{aligned}
		P(ni|\hat{s}&_{ni})=\frac{e^{\text{sim}(s_{ni},\hat{s}_{ni})/\tau}}{\sum\limits_{s}{e^{\text{sim}(s,\hat{s}_{ni})/\tau}}},\\ &s\in{S_{ni}^-}\cup S_{ni}^+.
	\end{aligned}
\end{equation}
On the other hand, the probability of incorrectly recognizing $\hat{s}_{ni}$ as $s_{n'i'} (n'\neq n\ or\ i'\neq i)$ can be formulated as:

\begin{equation}
	\begin{aligned}
		P(n'i'|& \hat{s}_{ni})=\frac{e^{\text{sim}(s_{n'i'},\hat{s}_{ni})/\tau}}{\sum\limits_{s}{e^{\text{sim}(s,\hat{s}_{ni})/\tau}}}.\\ 
	\end{aligned}
\end{equation}

Assuming each patch being recognized as $p_{ni}$ is independent, then the joint probability of making the right classifications can be formulated as:
\begin{equation}
	\begin{aligned}
		P_{ni}=&P(ni|\hat{s}_{ni})\prod_{n'\neq n\ or\ i'\neq i}{(1-P(n'i'|\hat{s}_{ni}))},\\
	\end{aligned}
\end{equation}

Patch discrimination attempts to optimize the model to make correct patch classification, i.e., to optimize $f(\cdot)$ and $g(\cdot)$ to maximize $P_{ni}$. We can apply minimizing the corresponding negative log-likelihood as the alternative training target, and the final loss function for patch discrimination ($\mathcal{L}_{Pd}$) can be formulated as:
\begin{equation}
	\label{equation:Pd}
	\mathcal{L}_{Pd}=\sum_{n,i}M_{ni}.
\end{equation}

\begin{equation}
	\begin{aligned}
		M_{ni}=-\text{log}(P_{ni}),
	\end{aligned}
\end{equation}
Through optimizing the model by minimizing $\mathcal{L}_{Pd}$, patch instances can be dispersedly projected over the learned hypersphere space and the cosine distance between embedded features can be used to measure the similarity between the correlated patches.

\subsection{Region Discrimination}
\label{subsection:region-discrimination}
\textcolor{R_P}{While patch discrimination is capable of extracting finer patch-level discriminative features, the mechanism of dividing an image instance into patches may impede the integrality of anatomical structures, and treating each patch as an individual class neglects the similarity among patches. Such a representation learning scheme has limited ability to capture the comprehensive structure information, and its training is easily disturbed by the patches with similar contents across different images since they are forcibly classified into different classes.}

\textcolor{R_P}{To address the aforementioned issues and satisfy the demand of medical image analysis for representation that is capable of reflecting comprehensive structure information and measuring pixel/region-wise similarity, we further propose region discrimination, which creatively incorporates the intra-modality structure similarity into the model design. More specifically, besides the embedding module for pixel-wise embedding, region discrimination is further equipped with the clustering module (shown in Sec.~\ref{subsection:backbone}) for generating segmentation to discriminate structure regions in an image and assimilate semantically consistent regions among different images. These two modules are jointly optimized by the novel region discrimination loss function, which enables the model to closely embed semantically similar pixels and distinguish regions of different semantic classes.}

As shown in Fig.~\ref{fig:framework}.(c) and Eq.~(\ref{equation:f_g_2_v_r}), the segmentation mask $r_n$ generated by the clustering module indicates the probability of classifying pixels into $C$ classes. \textcolor{R_P}{A high value of $r_{nc}(h,w)$ means the pixel $x_n(h,w)$ is categorized into $c$-th semantic class with high confidence. Therefore, when representing the $c$-th semantic region in $x_n$, the semantic information of the pixel with a high $r_{nc}$ is reasonable to play a more important role. Fusing this idea into the model design, we denote the embedded vector for $c$-th structure region in $x_n$ as $t_{nc}$ and it can be deduced by combining the pixel-wise embedding $v_n$ and the segmentation mask $r_{nc}$} as follows:
\begin{equation}
	\label{equation:t_nc}
	t_{nc}=\frac{\sum_{h,w}r_{nc}(h,w)v_n(h,w)}{\left\| \sum_{h,w}r_{nc}(h,w)v_n(h,w) \right\|_2}.
\end{equation} Since images of the same modality commonly share similar structure regions, we can obtain the prototype (cluster centroid) vector $T_c$ for the $c$-th semantic class as follows: 
\begin{equation}
	\label{equation:T_c}
	T_c=\frac{\sum_{n}t_{nc}}{\left\| \sum_{n}t_{nc} \right\|_2}.
\end{equation}
We thus get the embedded vector for each structure region and the prototype vector for each semantic class. Then, the optimization objective of region discrimination is to cluster regions of the same semantic class together, i.e., to assign $t_{nc}$ to the prototype of $T_c$. 

The probability of assigning $t_{nc}$ to $T_c$ can be defined as:

\begin{equation}
	\label{equation:region_clustering}
	\begin{aligned}
		P_{nc}=P(c|t_{nc})=\frac{e^{\text{sim}(T_c,t_{nc})/\tau}}{\sum\limits_{0\leq c'<C}e^{\text{sim}(T_{c'},t_{nc})/\tau}}.
	\end{aligned}
\end{equation}   
Since the negative log-likelihood of the joint probability $\prod \limits_{n,c} P_{nc}$ can be set as the loss function for region discrimination, we denote the loss as $\mathcal{L}_{Rd}$ and the detailed definition can be formulated as follows:
\begin{equation}
	\label{euqation:L_RD}
	\begin{aligned}
		&\ \mathcal{L}_{Rd}=-\sum_{n,c}\text{log}(P_{nc}),\\
	\end{aligned}
\end{equation} 
\textcolor{R_P}{
\noindent \textbf{Formulation analysis:} Minimizing $\mathcal{L}_{Rd}$ (Eq.~(\ref{euqation:L_RD})) equals to maximizing $P_{nc}$ (Eq.~(\ref{equation:region_clustering})). And Eq.~(\ref{equation:region_clustering}) can be re-written as:
\begin{equation}
	\label{equation:region_clustering_rewritten}
	\begin{aligned}
		P_{nc}=P(c|t_{nc})=\frac{1}{1+\sum\limits_{c'\neq c}   \frac{e^{\text{sim}(T_{c'},t_{nc})/\tau}}{e^{\text{sim}(T_c,t_{nc})/\tau}}}.
	\end{aligned}
\end{equation}  
Then, the training objective can be replaced by maximizing $\text{sim}(T_c,t_{nc})$ and minimizing $\text{sim}(T_{c'},t_{nc})$, i.e., pulling $t_{nc}$ toward $T_c$ and pushing $t_{nc}$ away from $T_{c'}$. To realize this target, according to the definition of $t_{nc}$ (Eq.~(\ref{equation:t_nc})), the pixel with the embedding $v_{n}(h,w)$ similar to $T_c$ should have a high probability value $r_{nc}(h,w)$, and vice versa. This mechanism enables the clustering module and the embedding module to be optimized in a mutually beneficial mechanism, where the clustering module generates $r_{nc}$ to identify which pixels should be closely embedded for the embedding module, and the embedding module generates $v_{n}$ to decide which pixels should be clustered together for the clustering module.}

\noindent \textcolor{R_P}{\textbf{The evolutionary optimization strategy:} As analyzed above, the optimization of the embedding module and the clustering module in region discrimination rely on each other. As a consequence, if directly optimizing these two modules from scratch, they will provide random information to disturb each other, which may lead to meaningless clustering or a training collapse, i.e., the clustering module maps all pixels into a single category. To address this issue, our representation learning framework is designed as an evolutionary model to ensure stable training, by which the embedding is firstly trained by patch discrimination to with the initial patch-wise discriminative ability and then the whole model is further optimized by region discrimination to possess the pixel-wise and region-wise discriminative ability.}

\subsection{Hypersphere Mixup}
\label{subsection:Hypersphere-Mixup}
Patch discrimination learning described in Sec.~\ref{subsection:patch-discrimination} mainly applies consistency regularization to the hypersphere, specifically, it gathers multiple views of the same patch together in the embedding space. This constraint encourages the proposed model to measure similarity in semantic context and makes patch instances sparsely distributed over the hyperspace. \textcolor{R_P}{However, since the number of training patch instances is limited, a large part of the embedding space is blank, that is, no patches are projected into it, which limits the generation of the learned representation \cite{zhang2018mixup,verma2019manifold}. To tackle this issue and constrain the learned representation with smoothness in space, we further propose hypersphere mixup to fill the void in between embedded vectors of path instances. Its main idea is to construct virtual training patches as the linear interpolations of real patch instances and these virtual patches should be mapped to the linear interpolations of the associated embedded vectors. By conducting hypersphere mixup, the hyperspace is smoother and the learned presentation is more generalized and robust.}

\textcolor{R_P}{In hypersphere mixup, we firstly construct a virtual training sample $\tilde{x}_{n_1n_2}$ as the linear interpolation of two real image samples ($x_{n_1}$ and $x_{n_2}$),} i.e.:
\begin{equation}
	\label{equation:image_mixup}
	\tilde{x}_{n_1n_2}=\lambda x_{n_1}+(1-\lambda) x_{n_2},\ \lambda \in {(0,1)}.
\end{equation}
After that, $\tilde{x}_{n_1n_2}$ is fed into the model to get pixel-wise embeddings $\tilde{v}_{n_1n_2}$, i.e., $\tilde{v}_{n_1n_2}=g(f(\tilde{x}_{n_1n_2}))$. Then, similar to Eq. (\ref{equation:patch_embedding}), the embedded vector for $i$-th virtual patch $\tilde{p}_{n_1n_2i}$ is denoted as $\tilde{s}_{n_1n_2i}$ and can be formulated as :
\begin{equation}
	\label{equation:image_mixup_features}
	\tilde{s}_{n_1n_2i}=\frac{\sum_{h\in[h_b,h_e),w\in[w_b,w_e)}\tilde{v}_{n_1n_2}(h,w)}{\left\|\sum_{h\in[h_b,h_e),w\in[w_b,w_e)}\tilde{v}_{n_1n_2}(h,w)\right\|_2}.\\
\end{equation}

On the other hand, we construct a virtual embedded vector $\overline{s}_{n_1n_2i}$, which is the linear interpolation of $s_{n_1i}$ and $s_{n_2i}$ with the same $\lambda$ as Eq.~(\ref{equation:image_mixup}), where $s_{n_1i}$ and $s_{n_2i}$ are respectively the embeddings for $i$-th patch in $x_{n_1}$ and $i$-th patch in $x_{n_2}$. The formulation is defined as follows:
\begin{equation}
	\label{equation:virtual_embedded_vector}
	\begin{aligned}
		\overline{s}_{n_1n_2i}=&\frac{\lambda s_{n_1i}+(1-\lambda) s_{n_2i}}{\left\|\lambda s_{n_1i}+(1-\lambda)\textsf{} s_{n_2i}\right\|_2} .
	\end{aligned}
\end{equation}

The training target for hypersphere mixup is to optimize $f(\cdot)$ and $g(\cdot)$ to make $\tilde{s}_{n_1n_2i}$ similar to $\overline{s}_{n_1n_2i}$. To realize this, we can also maximize the probability of recognizing $\tilde{s}_{n_1n_2i}$ as $\overline{s}_{n_1n_2i}$. Similar to the analysis of $\mathcal{L}_{Pd}$ (Sec.~\ref{subsection:patch-discrimination}), we set $\overline{S}_{n_1n_2i}^+=\{\overline{s}_{n_1n_2i}\}$ and $\overline{S}_{n_1n_2i}^-=\{\overline{s}_{n_1'n_2'i'}|n_1' \neq n_1\ or\ n_2' \neq n_2\ or\ i'\neq i\}$ as the positive embedding set and the negative embedding set for $\tilde{s}_{n_1n_2i}$. Accordingly, the probability of making right classification and the probability of making incorrect classification are respectively denoted as $P(n_1n_2i|\tilde{s}_{n_1n_2i})$ and $P(n_1'n_2'i'|\tilde{s}_{n_1n_2i})$ ($n_1' \neq n_1\ or\ n_2' \neq n_2\ or\ i'\neq i$), and their detailed formulations are as follows:

\begin{equation}
	\begin{aligned}
		&P(n_1n_2i|\tilde{s}_{n_1n_2i})=\frac{e^{\text{sim}(\overline{s}_{n_1n_2i},\tilde{s}_{n_1n_2i})/\tau}}{\sum\limits_{\overline{s}}{e^{\text{sim}(s,\tilde{s}_{n_1n_2i})/\tau}}},\\
		&P(n_1'n_2'i'|\tilde{s}_{n_1n_2i})=\frac{e^{\text{sim}(\overline{s}_{n_1'n_2'i},\tilde{s}_{n_1n_2i})/\tau}}{\sum\limits_{\overline{s}}{e^{\text{sim}(s,\tilde{s}_{n_1n_2i})/\tau}}},\\
		&\ \ \ \ \ \ \ \ \ \ \ \ \ \ \ \ \ \ \overline{s}\in {\overline{S}_{n_1n_2i}^-}\cup \overline{S}_{n_1n_2i}^+,\\
	\end{aligned}
\end{equation}
Then, the joint probability of making the correct classification can be defined as follows:
\begin{equation}
	\begin{aligned}
		P_{n_1n_2i}&=P(n_1n_2i|\tilde{s}_{n_1n_2i})\prod_{n_1',n_2',i'}{(1-P(n_1'n_2'i'|\tilde{s}_{n_1n_2i}))}.
	\end{aligned}
\end{equation}

\textcolor{R_P}{Finally, we can deduce the loss function for hypersphere mixup ($\mathcal{L}_{Hm}$) as the negative log-likelihood of the joint probability as: }
\begin{equation}
	\label{equation:hypersphere_mixup}
	\mathcal{L}_{Hm}=-\sum_{n_1,n_2,i}\text{log}(P_{n_1n_2i}).
\end{equation}

\section{Shape-guided Segmentation and One-shot Localization}
\label{section: Shape-guided Segmentation and One-shot Localization}
The previous section proposes a local discriminative representation learning framework composed of patch discrimination and region discrimination. Patch discrimination enables the trained model to measure similarity between patches, and region discrimination empowers the representation with the ability to cluster semantically similar pixels and distinguish regions of different semantic classes. In this section, we attempt to investigate its potential clinical applications. Specifically, in Sec.~\ref{subsection:Shape-guided-segmentation}, the shape prior of similar structures is firstly introduced to be fused with region discrimination to realize segmentation of the target structure with the expected shape distribution. Then, in Sec.~\ref{section:Center-sensitive One-shot localization}, we introduce the center-sensitive property into patch discrimination and the improved model is suitable for detecting landmarks with only one labeled sample. 

\subsection{Shape-guided Cross-modality Segmentation}
\label{subsection:Shape-guided-segmentation}
\begin{figure}[]
	\centering	
	\includegraphics[width=0.98\columnwidth]{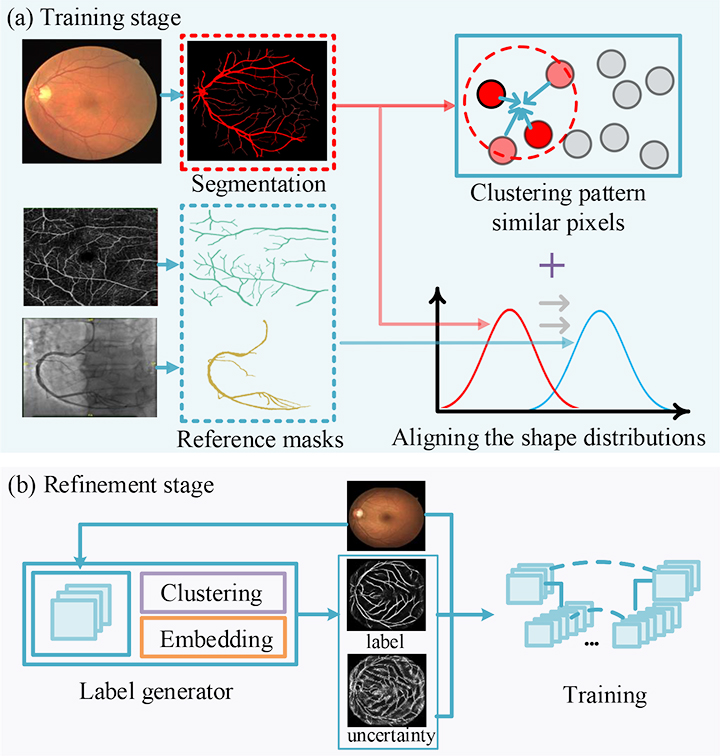}
	\caption{An illustration of the enhanced shape-guided cross-modality segmentation.} 
	\label{fig:shape-guided-segmentation}
\end{figure}

As inter-modality shape similarity indicated, some structures or tissues in different medical modalities, such as vessels in color fundus images, OCTA images, and DSA images, have similar shapes. If one of these structures has manual segmentation masks, it is practical and cost-effective to re-utilize the annotations to describe the shape distribution for another structure. Moreover, it is a common ability of clinicians to recognize the anatomical structure based on the shape description and its interior pattern consistency. Additionally, region discrimination can ensure the pixels clustered together share consistent patterns. Therefore, we attempt to demonstrate the feasibility of fusing the shape prior from similar structures with region discrimination to segment target structures. 

As shown in Fig.~\ref{fig:shape-guided-segmentation}.(a), in this framework, we harness region discrimination and the reference masks from similar structures to obtain the target region composed of pixels with similar patterns and with an expected shape. \textcolor{R_P}{Since KL divergence can measure the similarity between two distributions and the lower KL divergence the more similar corresponding distributions \cite{mackay2003information}, we optimize the model to minimize the Kullback-Leibler (KL) divergence between the probability distributions of the corresponding segmentation masks and reference masks to ensure the segmented structures with the expected shape. More specifically, we set the $m$-th segmentation mask ($\{r_{nm}\}_{n=1}^N$) as the target structure mask, and its probability distribution is denoted as $Ps_m$; meanwhile, we set the probability distribution of the reference masks as $Qs_m$, which is the reference probability distribution for $Ps_m$. Then, the optimization target is to update $f(\cdot)$ and $g(\cdot)$ to rectify the generated $r_{nm}$ to have similar probabilities in $Ps_m$ and $Qs_m$, i.e., $Ps_m(r_{nm})\approx Qs_m(r_{nm})$.} This training objective can be formulated as follows:

\begin{small}
	\begin{equation}
		\mathop{min}_{f(\cdot),g(\cdot)}KL(Ps_{m}||Qs_{m})=\sum_{n}Ps_{m}(r_{nm})\log\frac{Ps_{m}(r_{nm})}{Qs_{m}(r_{nm})}.
	\end{equation}
\end{small}

\textcolor{R_P}{Since adversarial learning is demonstrated to minimize a symmetrized and smooth version of KL divergence (Jensen-Shannon divergence) between two distributions \cite{goodfellow2014generative} and can be applied as an alternative optimizing strategy for minimizing KL divergence, we adopt it to align the distributions of segmentation masks and reference manual segmentation masks from other modalities.} The implementation details can be seen in Sec.~\ref{section:shape-guided_implementation}.

The initial segmentation model can be obtained as described above. However, this model is trained only with the guidance of reference masks and the constraint of region discrimination, the absence of definite annotations affects its robustness. Meanwhile, as presented in Sec.~\ref{subsection:region-discrimination}, the prototypes in region discrimination are deduced from all images, which neglects the specific characteristic of an individual image and hence hinders the segmentation from being adaptive for each test image. To tackle these issues, we propose an enhanced framework, containing re-clustering refinement and uncertainty estimation refinement, to get more accurate and robust results.

\noindent \textbf{Re-clustering refinement} is proposed to empower the segmentation to be adaptive to an individual test instance. Specifically, for an image instance $x_n$, we can not only get the initial segmentation result $r_{nm}$, but also get a clustered result based on pixel-wise embedding $v_n$ and the individual prototype vector $t_{nc}$ (formulated as Eq.~(\ref{equation:t_nc})). The re-clustered result $r_{nm}^*$ is defined as the probability of classifying pixels into the $m$-th individual prototype, i.e:
\begin{equation}
	\label{equation:refinement by clustering}
	r_{nm}^*(i,j)=\frac{e^{\text{sim}(t_{nm},v_n(h,w))/\tau}}{\sum \limits_{0\leq c<C}e^{\text{sim}(t_{nc},v_n(h,w))/\tau}}.
\end{equation}
The utilized prototype vector is specifically captured for the test sample, which better fits with the test instance and can improve the accuracy of segmentation.  

\noindent \textbf{Uncertainty estimation refinement} attempts to address the model's frangibility caused by the absence of definite annotations. Its working mechanism is to utilize uncertainty estimation \cite{kendall2017uncertainties,bian2020uncertainty} to identify reliable predictions as pseudo labels, and then re-train a segmentation model based on these labels in a supervised mechanism.

As shown in Fig.~\ref{fig:shape-guided-segmentation}.(b), pseudo labels and uncertainty maps are firstly generated by the label generator, which is composed of $f(\cdot)$ and $g(\cdot)$ and is trained by the constraint of region discrimination and shape prior. Specifically, we firstly utilize varied augmentations to process $x_n$ to generate multiple image samples, denote $x_{n}^e$ ($0\leq e<E$) as the $e$-th augmented sample. Then $x_{n}^e$ is fed into $f(\cdot)$ and $g(\cdot)$ to generate $r_{nm}^{*e}$ according to Eq.~(\ref{equation:refinement by clustering}). Based on the above processings, the pseudo label $r_{nm}^{**}$ and the uncertainty map $u_{nm}$ can be calculated as follows:

\begin{equation}
	\label{equation: pseudo label}
	r_{nm}^{**}(i,j)=\left\{
	\begin{aligned}
		&1,\ if \ \frac{1}{E} \sum \limits_{e} r_{nm}^{*e}(i,j)>0.5 \\
		&0,\ \text{otherwise}
	\end{aligned}
	\right.  
\end{equation}

\begin{equation}
	\label{equation: uncertainty map}
	\begin{aligned}
		u_{nm}(i,j)=&-\frac{1}{E}\sum\limits_{e}r_{nm}^{*e}(i,j)\text{log}(r_{nm}^{*e}(i,j))-\\
		&\frac{1}{E}\sum\limits_{e}(1-r_{nm}^{*e}(i,j)\text{log}(1-r_{nm}^{*e}(i,j))).
	\end{aligned}
\end{equation}

Finally, $(x_{nm},r_{nm}^{**},u_{nm})$ are utilized to re-train a segmentation network. We define the final out as $r'$ and the loss function is set as a binary cross-entropy loss weighted by the uncertainty map, the detailed definition is as follows:

\begin{equation}
	\label{equation:refinement loss}
	\begin{aligned}
		\mathcal{L}_{wbce}=&-\frac{1}{N}\sum_{n,i,j} u_{nm}(i,j) r_{nm}^{**}(i,j)\log r'_{nm}(i,j)-\\
		&-\frac{1}{N}\sum_{n,i,j} u_{nm}(i,j) (1-r_{nm}^{**}(i,j))\text{log}(1-r'_{nm}(i,j)).
	\end{aligned}
\end{equation}

\subsection{Center-sensitive One-shot Localization}
\label{section:Center-sensitive One-shot localization}

The ability to measure similarity is the mainstay of metric-learning-based one-shot methods\cite{snell2017prototypical}, and the trained model by patch discrimination is capable of measuring patch-wise similarity. Therefore, we attempt to take full advantage of patch discrimination to realize unsupervised one-shot landmark localization to alleviate the burden of manual annotating.

One-shot landmark localization aims to localize landmarks in unlabeled images (query samples) based on one labeled sample (support sample). A natural idea based on patch discrimination is that we can firstly construct a patch with the labeled point as the center and capture its embedded vector as the support embedded vector. Then, we calculate the cosine similarity between the support embedded vector and the vectors of patches in the query sample. After that, the detected point is identified as the center point of the best matching patch. However, when deducing the embedded vector for a patch instance, original patch discrimination gives equal importance to pixel-wise embeddings within it (shown in Eq.~\ref{equation:patch_embedding}), which makes the identification of the target point easily disturbed by its nearby pixels. 

To address this challenge, rather than utilizing averaging pooling to get patch embedded vectors, we present a center-sensitive averaging pooling, which makes the representation of a patch pay more attention to the patch's central regions and contributes to better-aligned results. The center-sensitive averaging pooling can be seen as a convolutional operation, and its kernel values get bigger when closer to the center point. The kernel weight $W_c \in{\mathbb{R}^{\lfloor \frac{H}{H_p} \rfloor \times \lfloor \frac{W}{W_p} \rfloor}}$ is defined as follows:
\begin{equation}
	\label{equation:kernel_weights}
	\begin{aligned}
		W_c&(h,w)=e^{-\frac{\text{dis}_h^2(h,w)+\text{dis}_w^2(h,w)}{2\sigma^2}},\\
		\text{dis}_h(h,w)=&\frac{h}{h_{cen}}-1,\ \text{dis}_w(h,w)=\frac{w}{w_{cen}}-1,\\
		h\in & [0,\lfloor H/H_p \rfloor), w\in [0,\lfloor W/W_p \rfloor),
	\end{aligned}
\end{equation} where $W_c(h,w)$ is a Gaussian-liked formulation and $\sigma$ is a hyper-parameter to control the level of focusing on the center region. $(h_{cen},w_{cen})$ is the center point of the kernel, i.e., $h_{cen}=0.5\lfloor H/H_p \rfloor$ and $w_{cen}=0.5 \lfloor W/W_p \rfloor$. 

\noindent \textbf{Training stage:} To make the trained model have the center-sensitive ability, we firstly fuse the center-sensitive averaging pooling into the formulation of patch-wise embedding $s_{ni}$ (Eq.~(\ref{equation:patch_embedding})) and $\overline{s}_{n_1n_2i}$ (Eq.~(\ref{equation:virtual_embedded_vector})) as:
\begin{equation}
	\begin{aligned}
		\dot{s}_{ni}&=\frac{\sum v_n(h,w)W_c(h-h_b,w-w_b)}{\left\|\sum v_n(h,w)W_c(h-h_b,w-w_b)\right\|_2},\\
		\dot{s}_{n_1n_2i}&=\frac{\sum\tilde{v}_{n_1n_2}(h,w)W_c(h-h_b,w-w_b)}{\left\|\ \sum\tilde{v}_{n_1n_2}(h,w)W_c(h-h_b,w-w_b)\right\|_2},\\
	\end{aligned}
\end{equation}
where $h\in[h_b,h_e)$ and $w\in[w_b,w_e)$. Then, $\dot{s}_{ni}$ and $\dot{s}_{n_1n_2i}$ are utilized to replace the original $s_{ni}$ and $\overline{s}_{n_1n_2i}$ to compute $\mathcal{L}_{Pd}$ (Eq.~(\ref{equation:Pd})) and $\mathcal{L}_{Hm}$ (Eq.~(\ref{equation:hypersphere_mixup})). Finally, model is trained based on the modified loss function, which ensures that the patch comparison focuses on the center region.

\begin{figure}[]
	\centering	
	\includegraphics[width=0.98\columnwidth]{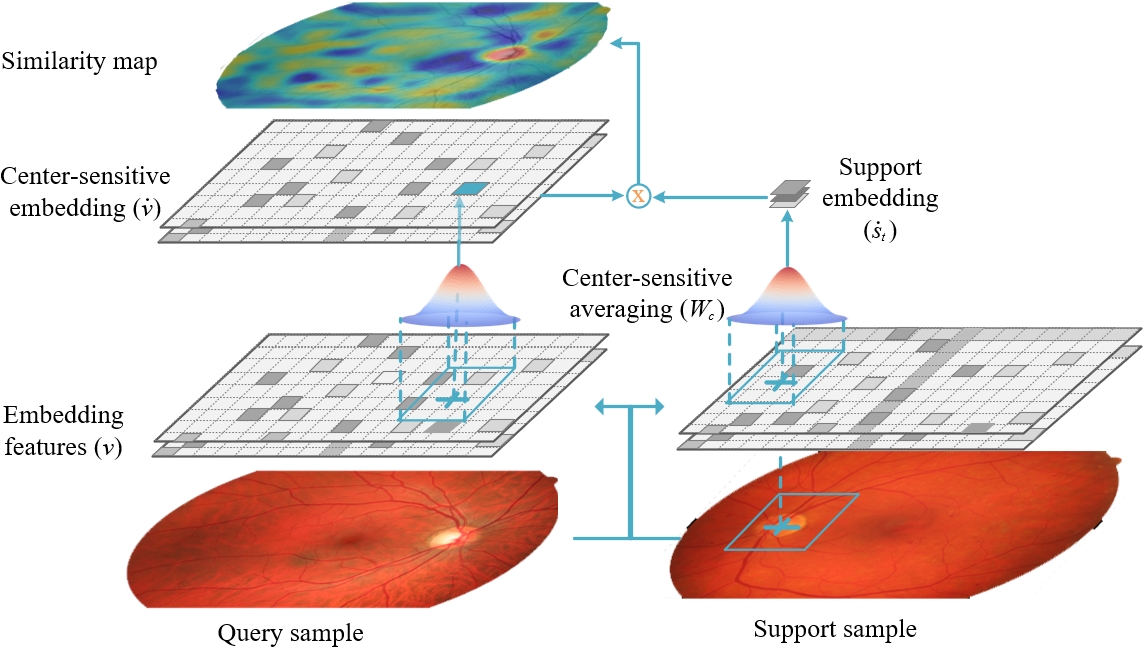}
	\caption{An illustration of the center-sensitive one-shot localization.} 
	\label{fig:one-shot-location}
\end{figure}

\noindent \textbf{Test stage:} As shown in Fig.~\ref{fig:one-shot-location}, the center-sensitive embedding for the test sample can be deduced by utilizing $W_c$ to filter original pixels-wise embedded vectors $v_n$, i.e.:
\begin{equation}
	\label{equation:one-shot-test}
	\dot{v}_n=conv(v_n,W_c).
\end{equation} 
Meanwhile, the support embedding $\dot{s}_t$ is set as the center-sensitive embedding of the patch with the labeled point as the center. Then, the similarity map are captured as the cosine distance between $\dot{s}_t$ and $\dot{v}_n$, i.e., $\text{sim}(\dot{s}_t,\dot v_n(h,w))=\dot{s}_t^\top \dot v_n(h,w)/(\left\|\dot{s}_t\right\|_2\left\|\dot v_n(h,w)\right\|_2)$. After that, the point with the max similarity value can be identified:
\begin{equation}
	(h_{max},w_{max})=\mathop{\text{argmax}}_{h,w} \ \text{sim}(\dot{s}_t,\dot v_n(h,w)).
\end{equation}
And we further adopt threshold segmentation to get regions with similarity values greater than 95\% of the max similarity value. Finally, we choose the connective region, which contains the point of $(h_{max},w_{max})$, and get its centroid point $(h_d,w_d)$ as the final result.

\section{Experimental Results}
\label{section: experiments}
This section contains three parts: (1) The generalization and robustness of our local discrimination model are analyzed in Sec.~\ref{section:Local_Discriminative_Representation_Learning}. (2) The comparative results of the shape-guided segmentation are presented in Sec.~\ref{section:Shape-guided Segmentation}. (3) Sec.~\ref{section:One-shot Localization} shows the results of the center-sensitive one-shot localization.

\subsection{Local Discriminative Representation Learning}
\label{section:Local_Discriminative_Representation_Learning}
\subsubsection{Datasets and Preprocessing}
\label{section:Datasets and preprocessing}
\begin{table}[]
	\caption{The list of datasets.}
	\centering
	\renewcommand\tabcolsep{2.2pt}
	\label{table:dataset list}
	
	\begin{threeparttable}
		\resizebox{\linewidth}{!}{
		\begin{tabular}{l|ccc}
			\toprule[0.5pt]
			\hline
			Name& Task & Train & Test\\	
			\hline	
			\hline
			\multicolumn{4}{c}{Color fundus datasets}\\		
			\hline
			Kaggle DR\tnote{1}\ \ \cite{cuadros2009eyepacs}& - &~35k&-\\ 
			CHASEDB (CHASE)\tnote{2}\ \ \cite{fraz2012ensemble}& vessel segmentation& 20 & 8\\
			HRF\tnote{3}\ \ \cite{budai2013robust}& vessel segmentation& 25 & 20\\
			
			RITE\tnote{4}\ \ \cite{hu2013automated}&vessel segmentation&20&20\\
			STARE\tnote{5}\ \ \cite{hoover2000locating,hoover2003locating}&vessel segmentation&10&10\\
			DRISHTI-GS1(Cup)\tnote{6}\ \ \cite{sivaswamy2015comprehensive,sivaswamy2014drishti}&optic cup segmentation&50&51\\ 
			DRISHTI-GS1(OD)\tnote{6}\ \ \cite{sivaswamy2015comprehensive,sivaswamy2014drishti}&optic disc segmentation&50&51\\
			IDRiD(OD)\tnote{7}\ \ \cite{h25w98-18}&optic disc segmentation&54&27\\  
			IDRiD(HE)\tnote{7}\ \ \cite{h25w98-18}&hard exudates segmentation&54&27\\
			\hline
			\multicolumn{4}{c}{Chest X-ray datasets}\\
			\hline
			ChestX-ray8\tnote{8}\ \ \cite{wang2017chestx}&-&~112k&-\\
			SCR(Lung)\tnote{9}\ \ \cite{van2006segmentation}&lung segmentation&124&123\\	
			SCR(Heart)\tnote{9}\ \ \cite{van2006segmentation}&heart segmentation&124&123\\	
			SCR(Clavicle)\tnote{9}\ \ \cite{van2006segmentation}&clavicle segmentation&124&123\\
			SIIM-ACR \tnote{10}&pneumothorax segmentation&1334&1335\\	
			\hline
		\end{tabular}
		}
		\begin{tablenotes}
			\footnotesize
			\item[1]https://www.kaggle.com/c/diabetic-retinopathy-detection/data
			\item[2]https://blogs.kingston.ac.uk/retinal/chasedb1/
			\item[3]https://www5.cs.fau.de/research/data/fundus-images/
			\item[4]https://medicine.uiowa.edu/eye/rite-dataset
			\item[5]http://cecas.clemson.edu/~ahoover/stare/
			\item[6]http://cvit.iiit.ac.in/projects/mip/drishti-gs/mip-dataset2/Home.php
			\item[7]https://idrid.grand-challenge.org/
			\item[8]https://nihcc.app.box.com/v/ChestXray-NIHCC
			\item[9]http://www.isi.uu.nl/Research/Databases/SCR/
			\item[10]https://www.kaggle.com/c/siim-acr-pneumothorax-segmentation
			
		\end{tablenotes}
	\end{threeparttable}
\end{table}

As shown in Table.~\ref{table:dataset list}, we evaluate our proposed method in color fundus datasets and chest X-ray datasets.

\noindent $\bullet$ \textbf{Color fundus datasets:} The color fundus images from Kaggle diabetic retinopathy (DR) detection dataset are firstly utilized to train the representation. Then, the trained representation is transferred to 8 downstream tasks.

\noindent $\bullet$ \textbf{Chest X-ray datasets:} The dataset of ChestX-ray8 with about $112k$ training data is utilized to train the representation. 4 downstream segmentation tasks are initialized by the learned representation.

For color fundus images, we firstly identify the field of view (FOV) by simple threshold segmentation and then crop images to filter out the background part. All of the above images are resized to $512\times 512$.

\subsubsection{Implementation Details}
\label{section: representation learning implementation details}
\textbf{Network architecture:} To make the training friendly in consumption time and GPU memory, we utilize a tiny version of VGG16 \cite{simonyan2014very} as the encoder $f(\cdot)$ and the decoder $g(\cdot)$ is also designed as a light-weighted network. Specifically, the setting of $f(\cdot)$ is almost the same as VGG16 without fully connected (FC) layers, but the number of channels is reduced to a quarter of VGG16. For $g(\cdot)$, each of the first four base blocks is composed of an up-pooling layer and two convolutional layers. In the final part of the decoder, the clustering module and the embedding module, each of which is made up of two convolutional layers, are set to obtain the segmentation mask $r_n$ and the pixel-wise embedding $v_n$ respectively. It is worth noting that the features of each encoder block are connected with the corresponding decoder block by the skip connection layer.

\noindent \textbf{Training process:} Learning local discriminative features is an evolutionary process from patch discrimination to region discrimination. Patch discrimination accompanied by hypersphere mixup is firstly utilized to train the model to get initial patch discriminative ability, the joint loss function can be formulated as:
\begin{equation}
	\mathcal{L}=\mathcal{L}_{Pd}+\mathcal{L}_{Hm}.
\end{equation}$f(\cdot)$ and $g(\cdot)$ are optimized by minimizing this loss in the first 20 epochs and updated $1k$ iterations in each epoch. 

After pre-training model based on patch-wise loss, region discrimination is further fused into the training process, and the total loss ($\mathcal{L}$) is defined as follows:
\begin{equation}
	\mathcal{L}=\mathcal{L}_{Pd}+\mathcal{L}_{Hm}+10\mathcal{L}_{Rd}+0.1\mathcal{L}_{entropy},
\end{equation}
\begin{footnotesize}
	\begin{equation}
		\begin{aligned}
			\mathcal{L}_{entropy}=&-\frac{1}{NCHW}\sum_{n,c,h,w}r_{nc}(h,w)\text{log}(r_{nc}(h,w))-\\
		&\frac{1}{NCHW}\sum_{n,c,h,w}(1-r_{nc}(h,w))\text{log}(1-r_{nc}(h,w)),
		\end{aligned}
	\end{equation}
\end{footnotesize}where $\mathcal{L}_{entropy}$ is a constraint on entropy to make $r_{nc}(h,w)$ with high confidence. The new loss is set to optimize the model in the following 80 epochs. 

In each training iteration, four data groups are fed into the model. Concretely, two images are firstly randomly sampled and mixed to generate the mixup image followed Eq.~(\ref{equation:image_mixup}); then, each sampled image is processed by different augmentation settings to generate two augmented versions; and the final group contains these four augmented images and the mixup image. The optimizer is an Adam with the initial learning rate ($lr$) of 0.001, and $lr$ will decline to half after every 10 epochs.

\noindent \textbf{Transfer to downstream tasks:} The segmentation model for downstream tasks is a U-Net, its encoder is the same as $f(\cdot)$ and its decoder is similar to $g(\cdot)$ but without the embedding module, and a final layer is activated by the Sigmoid. We firstly utilize the learned feature extractor as the initial encoder for models of downstream tasks and only update the decoder in the first 100 epochs. Then, all of the model's parameters are updated in the following 100 epochs. The Adam with $lr=0.001$ is utilized as the optimizer, and 20\% of the training data is split as the validation dataset to alleviate the over-fitting. The loss function is set as dice loss as follows:
\begin{equation}
	\begin{aligned}
		\mathcal{L}_{dice}=1&-2\frac{\sum_{n}y'_ny_n+\epsilon}{\sum_{n}y'_n+\sum_{n}y_n+\epsilon},\\
	\end{aligned}
\end{equation}where $y'_n \in Y',\  y_n \in Y$, $Y$ is the ground truth and $Y'$ is the prediction. The smoothness term $\epsilon$ is set as 1.

Since the number of the image in chest X-ray datasets (SCR and SIIM-ACR) is relatively abundant, we make comparative experiments to analyze the improvements brought by our method for different data numbers in Sec.~\ref{section:ablation experiments}. Specifically, we respectively use 20\%, 40\%, 60\%, 80\%, and 100\% of the training data to train the network. Accordingly, The evaluation metric for chest X-ray tasks in this section is the average metric value of results based on different ratios of the training data.

\noindent \textbf{Experiment setting:}
To construct augmentation image pairs, RandomResizedCrop, RandomGrayscale, ColorJitter,
RandomHorizontalFlip, and Rotation90 in PyTorch are adopted. $H_p$ and $W_p$ are set as 8 to divide an image into $8\times 8$ patches, and $C$ is set as 16 to cluster 16 regions for each image.

Experiments are performed on a workstation platform with Intel(R) Xeon(R) CPU E5-2640 v4 @ 2.40GHz, 128GB RAM and $4\times$ NVIDIA GeForce GTX 1080 GPU. The code is implemented in Ubuntu 18.04LTS with PyTorch 1.8.0.

\subsubsection{Comparative Experiments}
\begin{table*}[]
	\caption{Comparison of the results of downstream segmenting tasks (DSC\%).}
	\centering
	\renewcommand\tabcolsep{3.5pt}
	\label{table:comparative experiments}
	\begin{tabular}{l|cccc|ccc|c|c|ccc|c|c}
		\toprule[0.5pt]
		Modalities
		&\multicolumn{9}{|c|}{Color fundus}&
		\multicolumn{5}{|c}{Chest X-ray}\\
		\hline
		&\multicolumn{4}{|c|}{Retinal vessel}&
		\multicolumn{3}{|c|}{Optic disc and cup}
		&\multicolumn{1}{|c|}{Lesions}
		&\multicolumn{1}{|c|}{Mean}
		&\multicolumn{3}{|c|}{Anatomical structures}
		&\multicolumn{1}{|c|}{Lesions}
		&\multicolumn{1}{|c}{Mean}  \\
		\hline
		Methods& CHASE & HRF & RITE & STARE  & GS(cup) & GS(OD) & ID(OD)& ID(HE) &  & Heart & Lung & Clavicle & Pne\\				
		\hline
		\hline
		Scratch
		& 79.37 & 79.42 & 82.40  & 73.64 & 81.82 & 92.26 & 85.43 & 60.30 & 79.33 & 91.68 & 97.66 & 89.94 & 34.45 & 78.43\\
		
		Supervised
		& 79.51 & 79.68 & 82.84 & 80.36 & 87.78 & 95.98 & 93.75 & 61.27 & 82.64 & - & - & - & - & -\\
		\hline
		\hline
		
		\multicolumn{15}{c}{Contrastive learning based methods}\\
		\hline
		InstDisc \cite{wu2018unsupervised}
		& 75.08 & 75.81 & 79.44 & 73.33 & 82.87 & 91.41 & 90.34 & 56.44 & 78.09 & 92.51 & 97.63 & 89.96 & 31.14 & 77.81\\
		
		EmbInva \cite{ye2019unsupervised}
		& 79.61 & 79.23 & 81.23 & 78.50  & 83.94 & 96.73 & 94.30  & 58.53 & 81.50 & 90.98 & 97.76 & 90.60  & 35.77 & 78.77\\
		
		MoCo \cite{he2020momentum}
		& 79.48 & 79.91 & 81.62 & 76.41 & 86.46 & 95.75 & 95.23 & 64.09 & 82.36 & 93.78 & 97.88 & 92.39 & 37.72 & 80.44\\
		
		MoCoV2 \cite{chen2020improved}
		& 78.98 & 79.45 & 81.64 & 74.90  & 84.54 & 96.01 & 92.42 & 63.99 & 81.49 & 93.34 & 97.80  & 92.01 & 38.98 & 80.53\\
		
		SimCLR \cite{chen2020simple}
		& 79.10  & 79.83 & 82.93 & 78.78 & 84.77 & 95.47 & 92.87 & 58.99 & 81.59 & 93.75 & 98.01 & 92.32 & 38.76 & 80.71\\

		DenseCL \cite{wang2021dense}
		& 80.58 & 81.22 & 83.94 & 82.05 & 88.50  & 96.48 & 95.07 & 65.28 & 84.14 & 93.47 & 97.78 & 91.47 & 36.44 & 79.79 \\
		
		\hline
		\hline
		\multicolumn{15}{c}{Clustering-based methods}\\
		\hline
		
		\textcolor{R_P}{DC \cite{caron2018deep}}
		& 75.88 & 76.34 & 80.15 & 73.90  & 86.97 & 95.98 & 96.13 & 56.91 & 80.28 & 92.76 & 97.88 & 91.01 & 36.15 & 79.45 \\
		
		\textcolor{R_P}{JC\cite{chen2021jigsaw}}
		& 79.98 & 80.65 & 83.21 & 81.39 & \textbf{91.60}  & \textbf{97.41} & 93.92 & 61.88 & 83.75 & 94.00    & 97.92 & 92.72 & 40.28 & 81.23 \\
		
		\textcolor{R_P}{SwAV \cite{caron2020unsupervised}}
		& 80.08 & 80.61 & 83.44 & 80.28 & 85.94 & 96.98 & 90.78 & 63.63 & 82.71 & 93.67 & 97.91 & 91.70  & 40.84 & 81.03 \\
		
		\textcolor{R_P}{CC \cite{li2021contrastive}}
		& 80.33 & 81.00    & 83.83 & 80.47 & 88.87 & 96.95 & 95.16 & 62.54 & 83.64 & 93.76 & 97.93 & 92.16 & 40.44 & 81.07 \\
		
		\textcolor{R_P}{PCL \cite{li2020prototypical}}
		& 79.31 & 80.38 & 82.75 & 79.63 & 90.31 & 96.90  & 95.40  & 65.34 & 83.75 & 93.9  & \textbf{97.98} & 92.05 & 39.05 & 80.74 \\

		\hline
		\hline
		\multicolumn{15}{c}{Medical images methods}\\
		\hline
		\textcolor{R_P}{RotOri \cite{li2021rotation}}
		& 75.81 & 75.54 & 79.24 & 71.48 & 84.85 & 96.07 & 91.61 & 53.79 & 78.54 & 91.26 & 97.41 & 88.18 & 32.47 & 77.33\\
		
		\textcolor{R_P}{CR\cite{chen2019self}} 
		& 79.93 & 80.55 & 83.07 & 80.07 & 86.98 & 97.18 & 93.48 & 64.53 & 83.22 &91.80  & 97.81 & 91.66 & 37.28 & 79.63 \\
		
		\textcolor{R_P}{CGM\cite{biffi2018learning}} & 72.20  & 74.17 & 77.65 & 61.53 & 83.30  & 91.09 & 90.09 & 52.67 & 75.33 & 89.61 & 96.88 & 85.71 & 27.63 & 74.95 \\

		\textcolor{R_P}{C2L \cite{zhou2020comparing}} & 72.41 & 74.13 & 76.46 & 63.98 & 83.79 & 74.64 & 77.74 & 51.23 & 71.79 & 93.85 & 97.92 & 92.54 & 41.70  & 81.50 \\

		\textcolor{R_P}{SimTriplet \cite{liu2021simtriplet}} & 79.50  & 79.66 & 82.44 & 80.89 & 84.06 & 95.96 & 94.23 & 57.53 & 81.78& 93.64 & 97.94 & 92.46 & 38.96 & 80.75 \\

		\textcolor{R_P}{EMMD \cite{li2020self}} & 79.83 & 79.04 & 82.76 & 75.9  & 87.39 & 96.10  & 94.60  & 59.56 & 81.89 &-&-&-&-&-\\
		
		\hline
		\hline
		\multicolumn{15}{c}{Ours}\\
		\hline
		IPMI21\cite{chen2021unsupervised} &80.91 & 81.28 & 84.11 & 82.68 & 89.63 & 96.92 & 95.97 & 65.89 & 84.67 & 93.92 & 97.96 & 92.51 & 41.03 & 81.35 \\

		Pd
		& 80.80  & 81.37 & 83.74 & 82.21 & 87.43 & 95.72 & \textbf{96.59} & 65.66 & 84.19 & 93.19 & 97.71 & 90.34 & 38.70  & 79.98\\
		
		Pd+mix
		& 80.80  & 81.22 & 84.02 & 81.69 & 90.01 & 96.89 & 96.25 & 66.34 & 84.65 & 94.17 & 97.95 & 92.78 & 42.36 & 81.81\\
		
		Rd 
		& \textbf{81.09} & \textbf{81.40}  & \textbf{84.15} & \textbf{82.91} & 91.22 & 96.97 & 96.23 & \textbf{67.05} & \textbf{85.12} & \textbf{94.32} & 97.97 & \textbf{92.98} & \textbf{43.20}  & \textbf{82.11}\\
		\hline
		
	\end{tabular}
\end{table*}

We compare our representation learning framework with recent SOTA methods. The comparative methods contain: 

\noindent 1) \textit{Scratch} is the network trained from scratch. 

\noindent 2) \textit{Supervised} is the supervised representation learning framework based on classification labels.  

\noindent 3) \textit{Instance-wise contrastive learning}: InstDisc \cite{wu2018unsupervised}, EmbInva \cite{ye2019unsupervised}, MoCo \cite{he2020momentum}, MoCoV2 \cite{chen2020improved}, SimCLR \cite{chen2020simple}, DenseCL \cite{wang2021dense}. 

\noindent \textcolor{R_P}{4) \textit{Clustering-based methods}: {DC} \cite{caron2018deep}, 
{JC}\cite{chen2021jigsaw}, 
{SwAV} \cite{caron2020unsupervised}, 
{CC} \cite{li2021contrastive}, and
{PCL} \cite{li2020prototypical}. SwAV, CC, and PCL are cluster-wise contrastive learning methods.}

\noindent 5) \textcolor{R_P}{\textit{Self-supervised medical representation learning methods}: \textit{RotOri} \cite{li2021rotation}, 
\textit{CR}\cite{chen2019self}, 
\textit{CGM}\cite{biffi2018learning}, 
\textit{C2L} \cite{zhou2020comparing}, 
\textit{SimTriplet} \cite{liu2021simtriplet}, and 
\textit{EMMD} \cite{li2020self}.}

\noindent 6) \textit{IPMI21} \cite{chen2021unsupervised}, \textit{Pd}, \textit{Pd+mix}, and \textit{Rd} are respectively the model in the conference version, patch discrimination, patch discrimination accompanied by hypersphere mixup, and region discrimination.


The evaluation metric is Dice Similarity Coefficient (DSC): $DSC=\frac{2|Y\cap Y'|}{|Y|+|Y'|}$, where $Y'$ is the binary prediction and $Y$ is the ground truth. Table.~\ref{table:comparative experiments} shows the detailed comparative results, and Fig.~\ref{fig:compared_with_reference} shows the curve of normalized validation loss for IPMI21 and region discrimination. We can draw the following conclusions:
  
\begin{figure}[]
	\centering	
	\includegraphics[width=1\columnwidth]{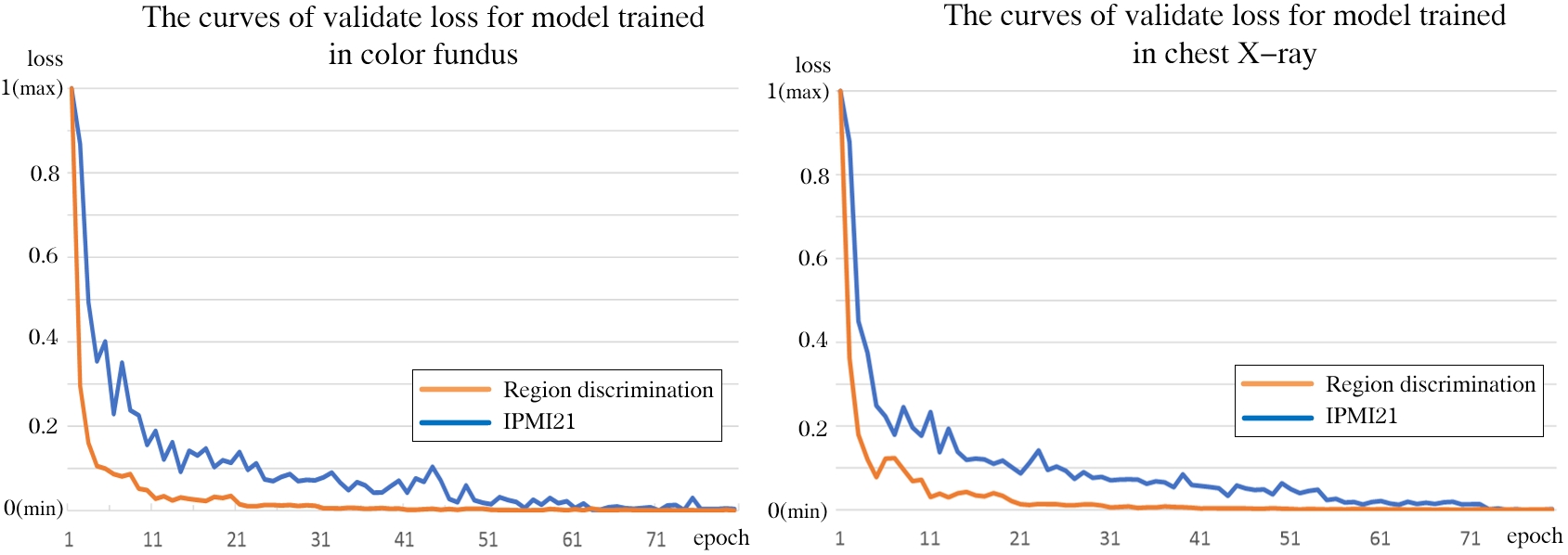}
	\caption{\textcolor{R_P}{The curve of normalized validation loss for IPMI21 and region discrimination. The x-coordinate shows the epoch number, and the y-coordinate shows the normalized loss, where 1 means the maximum value and 0 means the minimum value.} }
	\label{fig:compared_with_reference}
\end{figure}

1) Overall, the representation learned by our method is more generalized, which is evidenced by winning 9 tasks out of all 12 tasks. The pre-trained model has been proven for a better ability to depict local details and respectively gain improvements of $5.79\%$ and $3.68\%$ in the mean DSC for color fundus tasks and chest X-ray tasks.  

2) When compared with the supervised model, our method consistently achieves better performances in the downstream tasks. The reason is that the supervised model learns the condensed global features, which are not suitable to portray local details for medical tasks; meanwhile, these task-specific features may lack adaption and generalization and hence not be effective when being transferred for a different task\cite{8454781}.

\textcolor{R_P}{3) In comparison with SOTA instance-wise contrastive learning methods and clustering-based methods, our method achieves the best results. Most of these competitive methods measure inter-image similarity from a global perspective. Instead, our method focuses on pixel-wise clustering and region-wise comparisons, which generates finer features measuring local similarity for medical tasks.}

\textcolor{R_P}{4) Compared with recent SOTA methods for medical image analysis, our method demonstrates superior performances. The common medical property of intra-modality structure similarity enables our model to be adaptive to different medical application tasks; and in particular, our method excels for distinguishing tiny structures/tissues that are often considered as a common challenge to unsupervised models.}

\textcolor{R_P}{5) Compared to our pilot research \cite{chen2021unsupervised} (IPMI21), this new framework has better performances for multiple tasks on both color fundus images and chest X-ray images. In particular, our proposed method obtains significant improvements for lesion segmentation (a 1.16\% improvement for hard exudates segmentation and a 2.17\% improvement for pneumothorax segmentation). In IPMI21, the optimization objective is to minimize the distance between the pixel embedding and its assigned center embedding. Since the pixel-wise embedding is sensitive to the commonly existing noisy images with poor imaging conditions, the convergence of the model may experience dramatic fluctuations (shown in Fig.~\ref{fig:compared_with_reference}). Comparatively, the proposed region discrimination loss is based on the comprehensive and robust structure region embeddings, ensuring fast and stable convergence (shown in Fig.~\ref{fig:compared_with_reference}).}

6) Through comparing the results of Pd, Pd+mix, and Rd, we can see that our proposed evolutionary framework works well to obtain continuous improvements. Meanwhile, the effectiveness of hyperspace mixup (Pd+mix) is demonstrated by improving the mean DSC of patch discrimination by 0.46\% for color fundus tasks and 1.83\% for chest X-ray tasks.



\subsubsection{Ablation Experiments}
\label{section:ablation experiments}

In this part, ablation experiments are implemented to analyze the influences of \textit{patch-out size} (the setting of $(H_p,W_p)$), \textit{cluster number (the setting of $C$)}, and \textit{network complexity}. We also investigate the ability to \textit{reduce the demand for labeled data} of our method.

\begin{figure}[]
	\centering	
	\includegraphics[width=1\columnwidth]{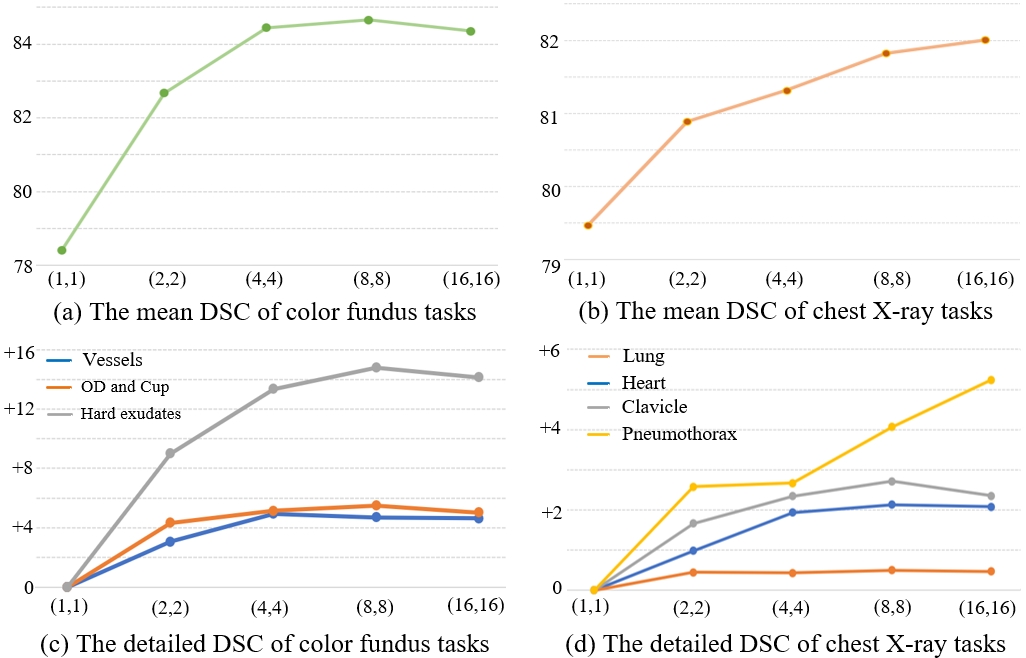}
	\caption{\textcolor{R_P}{The results of different patch-out sizes (DSC\%). (a) and (b) are respectively the mean DSC for color fundus tasks and chest X-ray tasks. (c) and (d) are respectively the detailed DSC improvements for different targets, where the y-coordinate means the improvement values compared to the result of the patch-out size with $(1,1)$.}} 
	\label{fig:ablation-patch-size}
\end{figure} 
\noindent \textbf{Patch-out size:} The model is trained by patch discrimination accompanied by hypersphere mixup, and the patch-out size $(H_p,W_p)$ is respectively set as $(1,1)$, $(2,2)$, $(4,4)$, $(8,8)$ and $(16,16)$. The mean DSC and the detailed DSC for different patch-out sizes are shown in Fig.~\ref{fig:ablation-patch-size}. We can see that:

1) The performance is rapidly improved when the patch-out size changes from $(1, 1)$ to $(8, 8)$, which shows the finer representation learned by comparing smaller patches is more generalized for downstream segmentation tasks. And the mean DSC changes slowly after the patch-out size is larger than $(8,8)$. Thus, we set $H_p=W_p=8$ in the following experiments.

\textcolor{R_P}{2) When the patch-out size changes from $(8,8)$ to $(16,16)$, the image is divided into more patches with smaller sizes. For the color fundus image with a large background region, the number of background patches lacking informative structure information significantly increases; consequently, the corresponding models are disturbed by these noisy patches, and the performance degenerates (shown in Fig.~\ref{fig:ablation-patch-size}(a)). For the chest X-ray image, it has more complex structures spreading over the whole image and with a relatively small background region, thus, the influence of background patches is less significant. Meanwhile, the comparisons among smaller patches contribute to features with a better ability to distinguish tiny regions, satisfying the demand of lesion segmentation. Therefore, the result of pneumothorax segmentation is significantly improved (shown in Fig.~\ref{fig:ablation-patch-size}.(d)), ensuring the total results go up (shown in Fig.~\ref{fig:ablation-patch-size}.(b)).}

\begin{figure}[]
	\centering	
	\includegraphics[width=1\columnwidth]{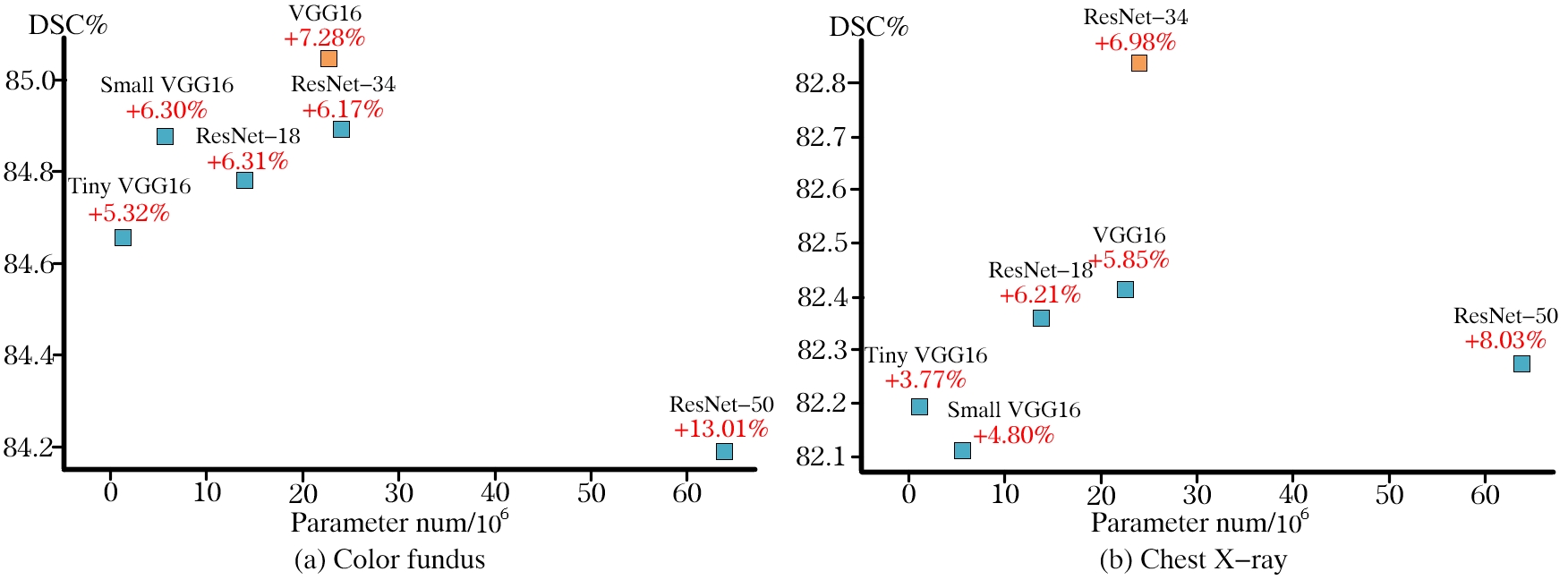}
	\caption{\textcolor{R_P}{The results of mean DSC for models with different complexities encoders (DSC\%). (a) and (b) are respectively for color fundus tasks and chest X-ray tasks, and the red font values mean the improvement values compared with the corresponding model trained from scratch.}}
	\label{fig:ablation-network-complexity}
\end{figure}
\begin{table}[b]
	\caption{\textcolor{R_P}{Parameter number for different models}}
	\centering
	\renewcommand\tabcolsep{3.5pt}
	\label{table:parameter_numbers}
	\begin{tabular}{c|ccc}
		\toprule[0.5pt]
		Encoder architecture&\multicolumn{3}{c}{Parameter number ($10^6$)}\\
		\hline
		&Encoder(pre-trained)&Decoder(scratch)&Total\\
		\hline
		Tiny VGG16 & 0.92  & 0.49  & 1.41 \\
		Small VGG16& 3.68  & 1.97  & 5.65 \\
		ResNet-18 & 11.2  & 2.66  & 13.86 \\
		VGG16& 14.72 & 7.86  & 22.58 \\
		ResNet-34 & 21.31 & 2.66  & 23.97 \\
		ResNet-50 & \textbf{23.54} & \textbf{40.37} & \textbf{63.91} \\
		\hline
	\end{tabular}
\end{table}
\begin{table*}[]
	\caption{Comparison of the results with different cluster numbers (DSC\%)}
	\centering
	\renewcommand\tabcolsep{3.5pt}
	\label{table:ablation experiments}
	\begin{tabular}{c|cccc|ccc|c|c|ccc|c|c}
		\toprule[0.5pt]
		Modalities
		&\multicolumn{9}{|c|}{Color fundus}&
		\multicolumn{5}{|c}{Chest X-ray}\\
		\hline
		&\multicolumn{4}{|c|}{Retinal vessel}&
		\multicolumn{3}{|c|}{Optic disc and cup}
		&\multicolumn{1}{|c|}{Lesions}
		&\multicolumn{1}{|c|}{Mean}
		&\multicolumn{3}{|c|}{Anatomical structures}
		&\multicolumn{1}{|c|}{Lesions}
		&\multicolumn{1}{|c}{Mean}  \\
		\hline
		Cluster num& CHASE & HRF & RITE & STARE  & GS(cup) & GS(OD) & ID(OD)& ID(HE) &  & Heart & Lung & Clavicle & Pne\\	
		
		\hline
		\hline
		0& 80.80  & 81.22 & 84.02 & 81.69 & 90.01 & 96.89 & 96.25 & 66.34 & 84.65 & 94.17 & 97.95 & 92.78 & 42.36 & 81.81\\
		4 & 80.58 & 81.18 & 83.97 & 82.73 & 89.14 & \textbf{97.21} & 95.24 & 66.06 & 84.51 & 94.15 & \textbf{97.97} & 92.86 & \textbf{44.34} & \textbf{82.33} \\
		8 & 81.06 & 81.31 & 84.02 & 82.86 & 89.82 & 97.10  & \textbf{96.36} & 66.87 & 84.92 & 94.14 & 97.95 & 92.75 & 43.66 & 82.12 \\
		12 & \textbf{81.23} & 81.15 & 83.56 & 82.05 & 90.56 & 96.99 & 96.25 & \textbf{67.12} & 84.86 & 94.26 & \textbf{97.97} & 92.72 & 43.67 & 82.15\\
		16 & 81.09 & \textbf{81.40}  & \textbf{84.15} & \textbf{82.91} & \textbf{91.22} & 96.97 & 96.23 & 67.05 & \textbf{85.12} & \textbf{94.32} & \textbf{97.97} & \textbf{92.98} & 43.20  & 82.11\\
		
		\hline
	\end{tabular}
\end{table*}
\noindent \textbf{Network complexity:} Encoders with different complexities are implemented as the feature extractor of our model, containing tiny VGG16 (similar to VGG16 but with a quarter of the channels), small VGG16 (similar to VGG16 but with half the channels), VGG16, \textcolor{R_P}{ResNet-18, ResNet-34, and ResNet-50}. And it is worth noting that the first layer of ResNet \cite{he2016deep}, which is originally a convolutional layer with the kernel of $7\times{7}$ and the stride of $2\times{2}$, is replaced by two $3\times{3}$ convolutional layers and a $2\times{2}$ max-pooling to provide finer features for segmentation. According to the results shown in Fig.~\ref{fig:ablation-network-complexity} and the detailed parameter information shown in Table.~\ref{table:parameter_numbers}, we have the following conclusions: 

1) Compared with models trained from scratch, models pre-trained by our method can get significant improvements (as the red font values shown in Fig.~\ref{fig:ablation-network-complexity}). 

\textcolor{R_P}{2) When the number of parameters in the randomly initialized decoder is relatively small, the performance of the model is improved with the increase of the model complexity. Specifically, all of tiny VGG16, small VGG16, ResNet-18, VGG16, and ResNet-34 have the corresponding decoder with no more than 8M (million) parameters. When the encoder changes from tiny VGG16 to small VGG16 to ResNet-18 to VGG16 to ResNet-34, the corresponding performance is roughly improved with the increase of model parameters. VGG16 and ResNet-34 have similar numbers of parameters, and they respectively gain the best results in color fundus tasks and chest X-ray tasks. }

\textcolor{R_P}{3) If the randomly initialized decoder has extremely large parameters, it is challenging to improve the performance with the increase of the model complexity. Specifically, the corresponding decoder for ResNet-50 has an extremely complex decoder with 40.37M parameters. Consequently, while ResNet-50 is the most complex model, its performance is not satisfactory.} 

\noindent \textbf{Cluster number:} We respectively set $C$ as 4, 8, 12, 16 and the comparative results are shown in Table.~\ref{table:ablation experiments}. We have the following conclusions: 

1) The models further trained by region discrimination have better results than models without it (i.e., $C=0$). The reason is that the model trained by region discrimination, focusing on clustering pixels to form regions with interior semantic consistency and distinguishing regions of different structures/tissues, can measure pixel-wise/region-wise similarity, satisfying the demand of medical image analysis for distinguishing tiny targets.

2) For a target with a small size, the model trained by region discrimination with a bigger $C$ can contribute to a better performance. For color fundus images, the target structures, such as vessels and lesions, are of small size, making results with $C=16$ the best. However, since targets in chest X-ray tasks have relatively big sizes, so it is not necessary to set a big clustering number and $C=4$ is enough to reach the best results.

\begin{figure}[]
	\centering	
	\includegraphics[width=0.98\columnwidth]{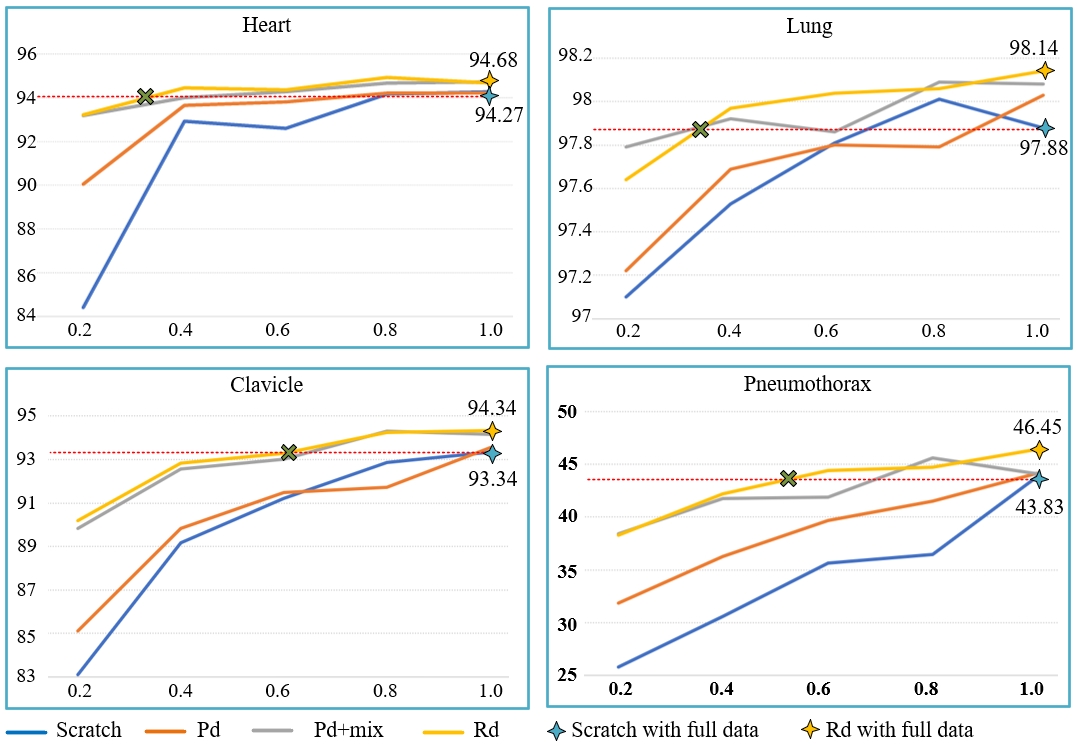}
	\caption{The results with different numbers of data (DSC\%).} 
	\label{fig:result-with-different-data-rate}
\end{figure}
\noindent \textbf{Annotation demand:} Based on the chest X-ray downstream tasks, we investigate the ability of our method to reduce the demand for labeled datasets. Specifically, 20\%, 40\%, 60\%, 80\%, and 100\% of the annotated data in the downstream tasks are set as the training data for models. According to the results shown in Fig.~\ref{fig:result-with-different-data-rate}, we can see that: 

1) Models based on the trained model of our method (Rd) have the best results in most situations (15 out of 20). 

2) In the situation that the training data is extremely scarce, i.e., only 20\% of training data are available, Rd can significantly improve the performances by 8.81\%, 7.1\%, and 12.49\% for the segmentation of heart, clavicle, and pneumothorax. 

3) Models based on Rd can quickly achieve better results with fewer data. Specifically, utilizing only 40\%, 40\%, 60\%, and 60\% of the training data can ensure that our models have better results than models trained from scratch with full data for the segmentation of heart, lung, clavicle, and pneumothorax.

\subsection{Shape-guided Segmentation}
\label{section:Shape-guided Segmentation}
\subsubsection{Implementation Details}
\label{section:shape-guided_implementation}
\textbf{Datasets:} About $11k$ color fundus images from Kaggle DR \cite{cuadros2009eyepacs} are set as the training data; 10 OCTA vessel masks and 10 DSA vessel masks are utilized as the reference masks. The trained model is evaluated in the test data of CHASEDB \cite{fraz2012ensemble}, HRF \cite{budai2013robust}, RITE \cite{hu2013automated} and STARE \cite{hoover2000locating,hoover2003locating}.

\noindent\textbf{Training details for clustering targets:}
Similar to Sec.~\ref{section: representation learning implementation details}, we firstly warm up models by patch discrimination accompanied by hypersphere mixup in the first 20 epochs. Then, in the following 80 epochs, we aim to generate segmentation masks ($R_{m}=\{r_{nm} \}_{n=1}^N$) to cluster target regions that are composed of semantically consistent pixels and possess a similar shape distribution to reference masks ($Re=\{re_n\}_{n=1}^N$). We utilize adversarial learning \cite{goodfellow2014generative} to align the probability distribution of predictions and reference masks. Specifically, besides the original $f(\cdot)$ and $g(\cdot)$, we build a discriminator network $d(\cdot)$ composed of 7 convolutional layers and 2 FC layers. The channel number for layers is respectively 16, 32, 32, 32, 32, 32, 32, 32, and 1. After each convolutional layer, a LeakyReLU and a $(2,2)$ max-pooling layer are set. Features of CNN are further processed by a global averaging pooling and fed into FC layers. The final out is activated by the Sigmoid. 

The training can be divided into two parts. In part one, the discriminator is updated to discriminate $Re$ and $R_{m}$. The loss function $\mathcal{L}_d$ is defined as:
\begin{equation}
	\mathcal{L}_d=\mathcal{L}_{bce}(d(Re),1)+\mathcal{L}_{bce}(d(R_m),0),
\end{equation} 
\begin{equation}
	\mathcal{L}_{bce}(y',y)=-\frac{1}{N}\sum\limits_{n}(y_n\text{log}y'_n+(1-y_n)\text{log}(1-y'_n)).
\end{equation}
In part two, $f(\cdot)$ and $g(\cdot)$ are updated to simultaneously cluster semantically similar pixels and cheat $d(\cdot)$. Accordingly, the loss function can be defined as follows:
\begin{equation}
	\mathcal{L}=\mathcal{L}_{Pd}+\mathcal{L}_{Hm}+10\mathcal{L}_{Rd}+0.1\mathcal{L}_{entropy}+\mathcal{L}_{adv},
\end{equation}
\begin{equation}
	\mathcal{L}_{adv}=\mathcal{L}_{bce}(d(R_m),1).
\end{equation}

The optimizers for both parts are Adams. And to ensure stable training, we set $lr$ to 0.00005 for the first part and 0.001 for the second part.

\noindent\textbf{Training details for refinement segmentation:}
We firstly capture 30 re-clustered results with different augmentations to identify pseudo labels and uncertainty maps according to Eq.~(\ref{equation: pseudo label}) and Eq.~(\ref{equation: uncertainty map}). Then a new network is built, which has similar architectures as $f(\cdot)$ and $g(\cdot)$, but the embedding module is removed from $g(\cdot)$ and the final layer of $g(\cdot)$ is a convolutional layer with only one output channel. The loss function is $\mathcal{L}_{wbce}$ defined in Eq.~(\ref{equation:refinement loss}). The optimizer is an Adam with $lr=0.001$ and the max training epoch is 100.

\noindent \textbf{Comparative methods:}
\textit{CycleGAN}: It is the basic model for synthetic segmentation \cite{dong2019synthetic,sandfort2019data}.
\textit{SynSeg} \cite{8494797}: It is an improved framework based on cycleGAN. \textit{IPMI21}: The results of the previous conference version. \textit{OutOri}: The results of the clustering module in our framework. \textit{OutReclu}: The results of re-clustering refinement. 
\textit{OutUncer}: The results of uncertainty estimation refinement.

\begin{table}[b]
	\caption{Comparison of the results of retinal vessel segmentation (DSC\%).}
	\centering
	\label{table:comparative DSA-guided}
	\begin{tabular}{l|cccc|c}
		\toprule[0.5pt]
		Methods& CHASEDB & HRF & RITE & STARE &Mean\\ 
		\hline
		&\multicolumn{5}{c}{DSA-guided segmentation}\\
		\hline
		\hline
		CycleGAN & 17.78 & 20.07 & 15.95 & 12.07 & 16.46 \\
		SynSeg & 35.61 & 24.31 & 50.36 & 48.04 & 39.58 \\
		IPMI21 &47.39 & 40.68 & 49.14 & 48.58 & 46.69\\
		OutOri & 57.82 & 52.52 & 64.71 & 62.64 & 59.42 \\
		OutReclu & 64.71 & 57.52 & 66.97 & 64.55 & 63.43 \\
		OutUncer & \textbf{68.22} & \textbf{59.96} & \textbf{70.3}  & \textbf{68.73} & \textbf{66.80} \\
		\hline
		&\multicolumn{5}{c}{OCTA-guided segmentation}\\
		\hline
		CycleGAN & 19.97 & 19.86 & 24.91 & 12.82 & 19.39 \\
		SynSeg & 11.75 & 24.11 & 22.51 & 28.97 & 21.83 \\
		IPMI21 &53.58 &52.27 &61.52 &58.74 & 56.53\\
		OutOri & 58.94 & 54.56 & 63.2  & 59.76 & 59.11 \\
		OutReclu & 61.41 & 57.59 & 63.26 & 61.91 & 61.04 \\
		OutUncer & \textbf{67.61} & \textbf{62.8}  & \textbf{66.02} & \textbf{66.97} & \textbf{65.85} \\
		\hline	
	\end{tabular}
\end{table}

\begin{figure}[t]
	\includegraphics[width=0.99\columnwidth]{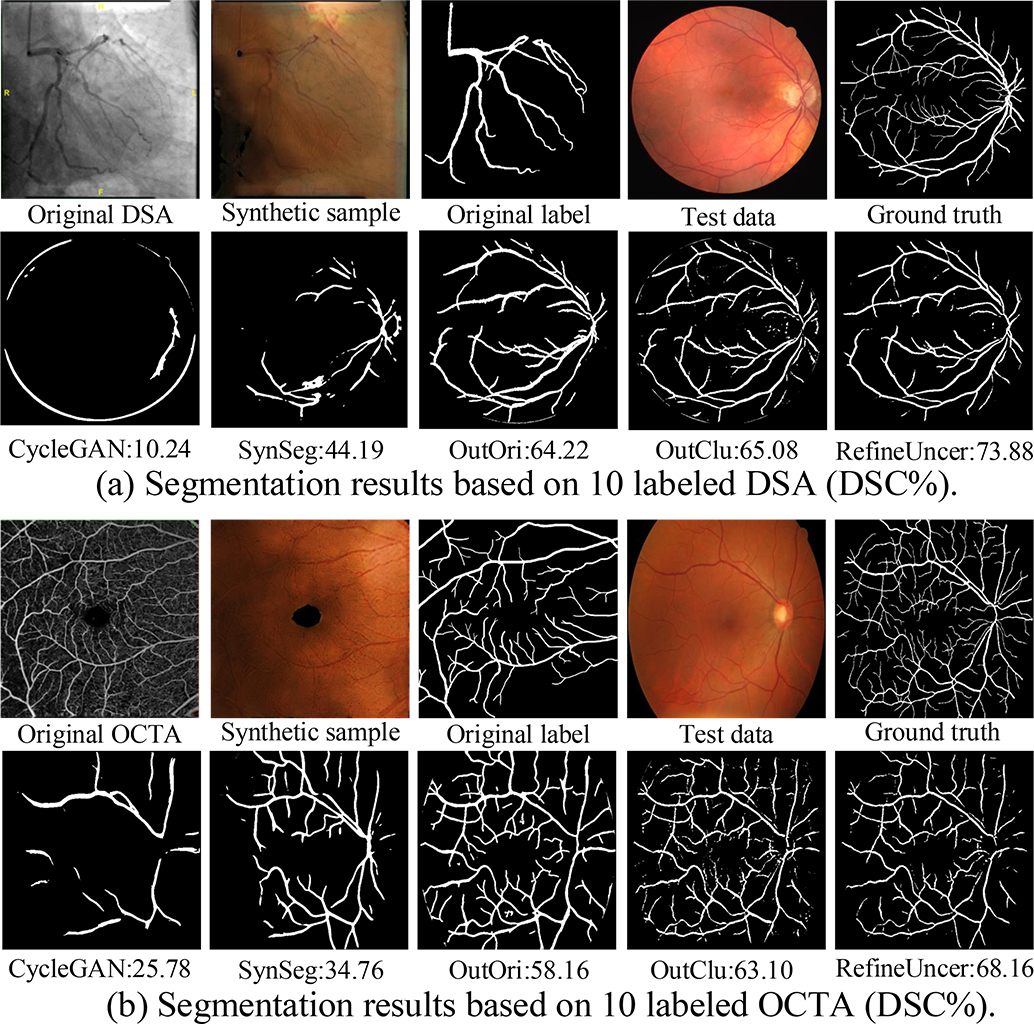}
	\caption{The visualization of segmentation results.} 
	\label{fig:shape-guided-segmentation-results}
\end{figure} 
\subsubsection{Experimental Results}
The results based on 10 DSA masks and 10 OCTA masks are shown in Table.~\ref{table:comparative DSA-guided}, and Fig.~\ref{fig:shape-guided-segmentation-results} shows some synthetic samples and predictions. The results demonstrate that:

1) Overall, when transferring knowledge across modalities with a big content gap, our method significantly outperforms other methods such as CycleGAN and SynSeg. Specifically, compared with SynSeg, our method gains a 27.22\% improvement in DSA-guided experiments, and a 44.02\% improvement in OCTA-guided experiments. As shown in Fig.~\ref{fig:shape-guided-segmentation-results}, although CycleGAN-based methods can get synthetic samples with similar styles as the color fundus image, the significant difference in contents makes the model trained based on the synthetic images failed in the real color fundus image.

2) Compared with our conference version (IPMI21), the enhanced model in this study (OutUncer) has significantly better results with 20.11\% DSC gains in DSA-guided segmentation and 9.32\% DSC gains in OCTA-guided segmentation, which shows its effectiveness and superiority.

3) In comparison with the results of OutOri, the results of OutReclu are better, demonstrating utilizing individual cluster prototypes to make pixel re-clustering for each image is effective. As shown in Fig.~\ref{fig:shape-guided-segmentation-results}, OutReclu can recognize more thin vessels compared with OutOri.

4) The results of OutUncer show that it is an effective strategy to further improve segmentation through utilizing reliable pseudo labels identified by the uncertainty estimation to re-train the model. Compared with OutClu, OutUncer can further remove false positives as shown in Fig.~\ref{fig:shape-guided-segmentation-results}.

\subsection{One-shot Localization}
\label{section:One-shot Localization}
\subsubsection{Implementation Details}

\textbf{Datasets:} 1) OD localization (HRF) \cite{budai2013robust}: The center of OD is manually labeled in 45 color fundus images. We utilize one data as the support data and the rest as the query data. 2) Heart localization (SCR) \cite{van2006segmentation}: We identify the centroid points based on the manual heart masks. We set one sample as the support data and 246 samples as the query data.

\noindent\textbf{Training details:} 
The basic model is trained based on patch discrimination accompanied by hypersphere mixup to obtain the ability to measure patches' similarity, specifically, the loss is defined as $\mathcal{L}=\mathcal{L}_{Pd}+\mathcal{L}_{Hm}$. Then, we conduct the following two improvements to make the learned representation more suitable for landmark localization:

1) Multi-size pooling: Rather than utilizing the adaptive average pooling with a fixed output size to gain the patch-wise representation, we use the average pooling layer with a random pooling size ranging from 28 to 112 to ensure that our model is adaptive for targets with varied sizes.

2) Center-sensitive averaging: As described in Sec.~\ref{section:Center-sensitive One-shot localization}, the averaging pooling layer is replaced by the center-sensitive averaging pooling to obtain well-aligned predictions. The $\sigma$ in Eq.~(\ref{equation:kernel_weights}) is set as 0.5. 

The total training epoch is 100, the optimizer is an Adam with $lr=0.001$, and four groups of images form a batch as Sec.~\ref{section: representation learning implementation details}.

\noindent \textbf{Evaluation metric:} To evaluate the performance, we firstly compute the euclidean distance between the real landmark point $(h_t,w_t)$ and the detected point $(h_d,w_d)$ as follows:
\begin{equation}
	\text{dis}((h_d,w_d),(h_t,w_t))=\sqrt{(h_d-h_t)^2+(w_d-w_t)^2}.
\end{equation}
Then, we define various threshold distances to identify the correct detection; specifically, the landmark is correctly localized if the distance between it and the predicted point is less than the threshold distance. The threshold distance is respectively set as 0.05, 0.1, 0.15, 0.2, 0.25 of the pooling size.

\noindent \textcolor{R_P}{\textbf{Competitive methods:} In addition to the comparison with (1) 2 recent self-supervised one-shot landmark localization methods (CC2D \cite{yao2021one}, RPR-Loc \cite{lei2021contrastive}), we compare our method with 9 template matching methods, which are also based on a support sample for landmark localization, including (2) 5 template matching methods based on hand-crafted features (CCOEFF-N \cite{bradski2008learning}, CCORR \cite{bradski2008learning}, CCORR-N \cite{bradski2008learning}, SQDIFF \cite{bradski2008learning}, SQDIFF-N \cite{bradski2008learning}), and (3) 4 template matching methods based on the representation learned by SOTA contrastive representation learning (MoCo \cite{he2020momentum}, C2L \cite{zhou2020comparing}, SimTriplet \cite{liu2021simtriplet}, DenseCL \cite{wang2021dense}).}

\begin{table}[]
	\caption{\textcolor{R_P}{Accurate rates with different threshold distances for one-shot localization (\%).}}
	\centering
	\label{table:one shot localization_comparative}
	\begin{tabular}{l|ccccc|c}
		\toprule[0.5pt]
		Methods& 0.05 & 0.1 & 0.15 & 0.2 &0.25 & Mean\\ 
		\hline
		\hline
		\multicolumn{7}{c}{OD localization}\\
		\hline
		CCOEFF-N & 22.72 & 40.90 & 40.90 & 43.18 & 43.18 & 38.18 \\
		CCORR & 4.54 & 18.18 & 31.81 & 36.36 & 45.45 & 27.27 \\
		CCORR-N & 13.63 & 31.81 & 31.81 & 31.81 & 31.81 & 28.18 \\
		SQDIFF & 15.90 & 34.09 & 34.09 & 34.09 & 34.09 & 30.45 \\
		SQDIFF-N & 18.18 & 34.09 & 34.09 & 36.36 & 36.36 & 31.81\\
		\hline
		MoCo & 4.54 & 15.90 & 20.45 & 27.27 & 27.27 & 19.09 \\
		C2L & 4.54 & 6.81 & 13.63 & 20.45 & 22.72 & 13.63 \\
		SimTriplet & 9.09 & 25    & 31.81 & 43.18& 54.54 & 32.72 \\
		DenseCL & 6.81 & 18.18 & 38.63 & 54.54 & 63.63 & 36.36 \\
		\hline
		CC2D & 0.00  & 6.82  & 22.73  & 29.55  & 29.55  & 17.73  \\
		RPR-Loc & 0.00 &2.27&2.27&11.36 &18.18 &6.82 \\
		\hline
		Ours & \textbf{25} & \textbf{61.36} & \textbf{88.63} & \textbf{97.72} & \textbf{97.72} & \textbf{74.09} \\

		\hline
		\hline
		\multicolumn{7}{c}{Heart localization}\\
		\hline
		CCOEFF-N & 2.44  & 14.23  & 26.02  & 37.80  & 46.34  & 25.37  \\
		CCORR & 0.41  & 0.41  & 0.41  & 0.41  & 0.41  & 0.41  \\
		CCORR-N & 2.44  & 12.60  & 28.46  & 38.62  & 54.88  & 27.40  \\
		SQDIFF & 1.63  & 6.10  & 14.23  & 20.33  & 27.24  & 13.90  \\
		SQDIFF-N & 1.63  & 6.10  & 13.82  & 20.33  & 27.64  & 13.90  \\
		\hline
		MoCo & 2.44  & 8.13  & 13.41  & 19.51  & 27.24  & 14.15  \\
		C2L & 3.66  & 10.57  & 18.29  & 29.27  & 37.80  & 19.92  \\
		SimTriplet & 2.44  & 4.47  & 9.35  & 15.04  & 19.92  & 10.24  \\
		DenseCL & 5.28  & 19.11  & 34.15  & 52.03  & 65.85  & 35.28  \\
		\hline
		CC2D  & 8.54  & 26.02  & 49.19  & \textbf{67.48}  & 80.49  & 46.34  \\
		RPR-Loc  & 6.10  & 22.36  & 42.68  & 62.20  & \textbf{80.89}  & 42.85  \\

		\hline
		Ours & \textbf{18.29}  & \textbf{44.31}  & \textbf{55.28}  & 63.82  & 67.07  & \textbf{49.76}  \\

		\hline	
	\end{tabular}
\end{table}

\begin{table}[]
	\caption{Ablation experiments for one-shot localization (\%)}
	\centering
	\label{table:one shot localization_Ablation}
	\begin{tabular}{l|ccccc|c}
		\toprule[0.5pt]
		Methods& 0.05 & 0.1 & 0.15 & 0.2 &0.25 & Mean\\ 
		\hline
		\hline
		\multicolumn{7}{c}{OD localization}\\
		\hline
		Baseline & 6.81 & 22.72 & 36.36 & 54.54 & 65.90 & 37.27 \\
		Baseline+ms & 6.81 & 29.54 & 59.09 & 72.72 & 88.63 & 51.36 \\
		Baseline+ms+cs & \textbf{25} & \textbf{61.36} & \textbf{88.63} & \textbf{97.72} & \textbf{97.72} & \textbf{74.09} \\

		\hline
		\multicolumn{7}{c}{Heart localization}\\
		\hline
		Baseline & 2.03  & 7.32  & 17.89  & 32.93  & 46.75  & 21.38  \\
		Baseline+ms & 1.22  & 10.98  & 23.58  & 36.18  & 52.85  & 24.96  \\
		Baseline+ms+cs & \textbf{18.29}  & \textbf{44.31}  & \textbf{55.28}  & \textbf{63.82}  & \textbf{67.07}  & \textbf{49.76}  \\
		\hline	
	\end{tabular}
\end{table}
 
\subsubsection{Experimental Results} 
The comparative results are shown in Table.~\ref{table:one shot localization_comparative} and the results of the ablation experiments are shown in Table.~\ref{table:one shot localization_Ablation}, where \textit{Baseline} is the original patch discrimination accompanied by hypersphere mixup, \textit{Baseline+ms} is the enhanced model with multi-size pooling, and \textit{Baseline+ms+cs} is the improved model with multi-size center-sensitive averaging pooling. Furthermore, some localization results and similarity maps of Baseline+ms+cs are shown in Fig.~\ref{fig:one-shot-localization-results}. Accordingly, we come to the following conclusions:

\begin{figure}[]
	\includegraphics[width=0.99\columnwidth]{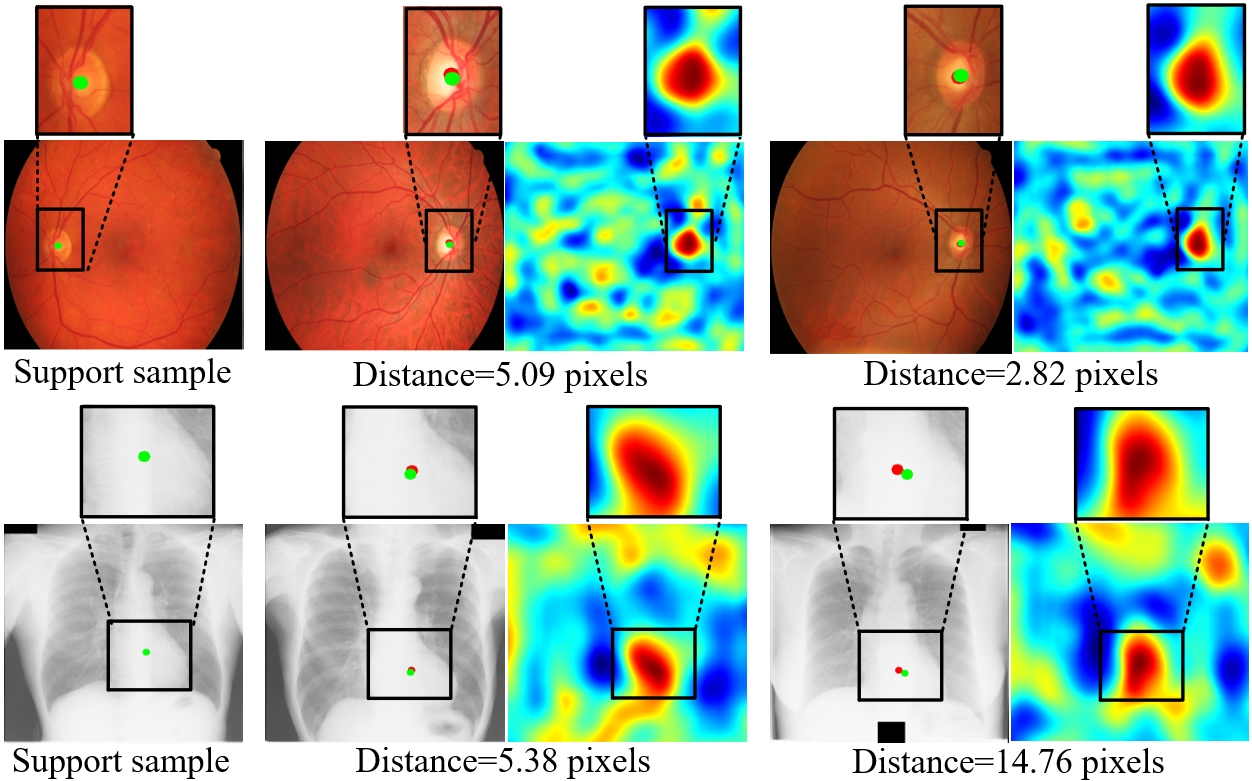}
	\caption{Some visualization samples of results for our one-shot localization method.} 
	\label{fig:one-shot-localization-results}
\end{figure} 

1) Overall, our proposed center-sensitive one-shot landmark localization method is more generalized and gains the best results in both tasks.

\textcolor{R_P}{2) For template matching methods, reliable features that are capable of measuring regional similarity are critically important to achieving stable performance in landmark localization tasks. Hand-crafted features are sensitive to image characteristics, such as color and contrast, leading to unstable performances in different tasks (as shown by the results of CCORR, SQDIFF, and SQDIFF-N). Since most of the SOTA contrastive learning methods, such as MoCo and SimTriplet, aim to learn global discriminative features, neglecting regional similarity, the representation learned by these methods may not be suitable for the localization tasks. Comparatively, the focus on measuring patch-wise similarity and the design of center-sensitive averaging pooling, secure our model the robust performance in achieving accurate landmark localization.}

\textcolor{R_P}{3) While the self-supervised one-shot localization methods (CC2D, RPR-Loc) achieve relatively competitive results in the heart localization task for chest X-ray images, the performance of both methods declines dramatically in the OD localization task for color fundus images. The major reason for the unsatisfactory results of CC2D is that the CC2D is based on a pixel matching mechanism, which is fragile for the color fundus image with a large number of meaningless background pixels. Since RPR-Loc requires the training dataset with stable relative positions, the differences between the left and the right eyes, the influence of different photography centers, and the varied fields of view, all together account for the less satisfactory results from RPR-Loc on color fundus images. In comparison, the augmentation-invariant and orientation-invariant patch discriminative features as learned and utilized in our method contribute to better generalization and robust performance in both medical tasks.}

4) Further, as shown in Fig.~\ref{fig:one-shot-localization-results}, the center-sensitive averaging pooling empowers the model to learn features focusing on the central region of a patch. Thus, the matching point has a peak value in the heat-map and the detected results align well with the target localization (shown by the results of threshold distance with 0.05 pooling size in Table.~\ref{table:one shot localization_comparative}).

5) As shown by the results in Table.~\ref{table:one shot localization_Ablation}, the effectiveness of the multi-size pooling and the center-sensitive averaging pooling are evidenced by the continuous improvements.

\section{Conclusion}
\label{section:conclusion}

In this work, we proposed an unsupervised local discriminative representation learning framework to address the challenging task of distinguishing tiny structures and tissues from different medical modality images and for various medical applications. The generalization capacity of this model is attributed to the common property of intra-modality structure similarity. Novel region discrimination is proposed to jointly train a clustering module to generate segmentation and an embedding module to closely embed semantically consistent pixels, which hence enables the representation to reflect more comprehensive structure information as well as to measure pixel/region-wise similarity. Furthermore, the training stability is enhanced with the evolutionary optimization strategy to pre-train the embedding with the initial patch-wise discriminative ability by the proposed patch discrimination accompanied by hypersphere mixup. 

The generalization and robustness of the representation learned by our method are fully validated on 12 downstream tasks, and our method achieves the best results on 9 tasks when compared to 18 SOTA methods. Based on the thorough analyses of the clinical applications, our framework attains a new perspective on knowledge transfer among modalities with enormous content gaps exemplar with vessel segmentation in color fundus images taking only shape prior from different imaging modalities of OCTA and DSA. Moreover, patch discrimination is further enhanced by the center-sensitive averaging pooling for one-shot landmark localization, which facilitates the detection of OD center and heart center based on one labeled data.  

\textcolor{R_P}{Discriminative representation learning for medical image analysis remains an open question, and there are three particular challenges to be addressed in our future work. Firstly, while our method has comprehensively explored the intra-modality structure similarity and the inter-modality shape similarity to learn discriminative features and transfer shape knowledge to realize segmentation, further intra-modality properties could be taken into account to learn more generalized discriminative features. Secondly, an adaptive algorithm, which may be designed as a meta-learning model to identify the hyperparameters, could further improve our model. Finally, we would like to translate the open-source framework to potentially support more clinical applications, such as one-shot medical image segmentation and anomaly detection.}

\appendices

\ifCLASSOPTIONcaptionsoff
  \newpage
\fi

\bibliographystyle{IEEEbib}
\bibliography{mybibfile}

\begin{thebibliography}{10}

\bibitem{ahmed2020artificial}
Zeeshan Ahmed, Khalid Mohamed, Saman Zeeshan, and XinQi Dong,
\newblock ``Artificial intelligence with multi-functional machine learning
  platform development for better healthcare and precision medicine,''
\newblock {\em Database}, vol. 2020, 2020.

\bibitem{meyer2018survey}
Philippe Meyer, Vincent Noblet, Christophe Mazzara, and Alex Lallement,
\newblock ``Survey on deep learning for radiotherapy,''
\newblock {\em Computers in biology and medicine}, vol. 98, pp. 126--146, 2018.

\bibitem{ardila2019end}
Diego Ardila, Atilla~P Kiraly, Sujeeth Bharadwaj, Bokyung Choi, Joshua~J
  Reicher, Lily Peng, Daniel Tse, Mozziyar Etemadi, Wenxing Ye, Greg Corrado,
  et~al.,
\newblock ``End-to-end lung cancer screening with three-dimensional deep
  learning on low-dose chest computed tomography,''
\newblock {\em Nature medicine}, vol. 25, no. 6, pp. 954--961, 2019.

\bibitem{litjens2017survey}
Geert Litjens, Thijs Kooi, Babak~Ehteshami Bejnordi, Arnaud Arindra~Adiyoso
  Setio, Francesco Ciompi, Mohsen Ghafoorian, Jeroen~Awm Van Der~Laak, Bram
  Van~Ginneken, and Clara~I S{\'a}nchez,
\newblock ``A survey on deep learning in medical image analysis,''
\newblock {\em Medical image analysis}, vol. 42, pp. 60--88, 2017.

\bibitem{chen2019self}
Liang Chen, Paul Bentley, Kensaku Mori, Kazunari Misawa, Michitaka Fujiwara,
  and Daniel Rueckert,
\newblock ``Self-supervised learning for medical image analysis using image
  context restoration,''
\newblock {\em Medical image analysis}, vol. 58, pp. 101539, 2019.

\bibitem{zhu2020rubik}
Jiuwen Zhu, Yuexiang Li, Yifan Hu, Kai Ma, S~Kevin Zhou, and Yefeng Zheng,
\newblock ``Rubik’s cube+: A self-supervised feature learning framework for
  3d medical image analysis,''
\newblock {\em Medical image analysis}, vol. 64, pp. 101746, 2020.

\bibitem{li2020self}
Xiaomeng Li, Mengyu Jia, Md~Tauhidul Islam, Lequan Yu, and Lei Xing,
\newblock ``Self-supervised feature learning via exploiting multi-modal data
  for retinal disease diagnosis,''
\newblock {\em IEEE Transactions on Medical Imaging}, vol. 39, no. 12, pp.
  4023--4033, 2020.

\bibitem{li2021rotation}
Xiaomeng Li, Xiaowei Hu, Xiaojuan Qi, Lequan Yu, Wei Zhao, Pheng-Ann Heng, and
  Lei Xing,
\newblock ``Rotation-oriented collaborative self-supervised learning for
  retinal disease diagnosis,''
\newblock {\em IEEE Transactions on Medical Imaging}, vol. 40, no. 9, pp.
  2284--2294, 2021.

\bibitem{han2021pneumonia}
Yan Han, Chongyan Chen, Ahmed Tewfik, Ying Ding, and Yifan Peng,
\newblock ``Pneumonia detection on chest x-ray using radiomic features and
  contrastive learning,''
\newblock in {\em IEEE 18th International Symposium on Biomedical Imaging
  (ISBI)}. IEEE, 2021, pp. 247--251.

\bibitem{zhang2020contrastive}
Yuhao Zhang, Hang Jiang, Yasuhide Miura, Christopher~D Manning, and Curtis~P
  Langlotz,
\newblock ``Contrastive learning of medical visual representations from paired
  images and text,''
\newblock {\em arXiv preprint arXiv:2010.00747}, 2020.

\bibitem{chaitanya2020contrastive}
Krishna Chaitanya, Ertunc Erdil, Neerav Karani, and Ender Konukoglu,
\newblock ``Contrastive learning of global and local features for medical image
  segmentation with limited annotations,''
\newblock {\em Advances in Neural Information Processing Systems}, vol. 33, pp.
  12546--12558, 2020.

\bibitem{sowrirajan2021moco}
Hari Sowrirajan, Jingbo Yang, Andrew~Y Ng, and Pranav Rajpurkar,
\newblock ``Moco pretraining improves representation and transferability of
  chest x-ray models,''
\newblock in {\em Medical Imaging with Deep Learning}. PMLR, 2021, pp.
  728--744.

\bibitem{tortora2018principles}
Gerard~J Tortora and Bryan~H Derrickson,
\newblock {\em Principles of anatomy and physiology},
\newblock John Wiley \& Sons, 2018.

\bibitem{chen2021unsupervised}
Huai Chen, Jieyu Li, Renzhen Wang, Yijie Huang, Fanrui Meng, Deyu Meng, Qing
  Peng, and Lisheng Wang,
\newblock ``Unsupervised learning of local discriminative representation for
  medical images,''
\newblock in {\em International Conference on Information Processing in Medical
  Imaging}. Springer, 2021, pp. 373--385.

\bibitem{he2020momentum}
Kaiming He, Haoqi Fan, Yuxin Wu, Saining Xie, and Ross Girshick,
\newblock ``Momentum contrast for unsupervised visual representation
  learning,''
\newblock in {\em Proceedings of the IEEE/CVF Conference on Computer Vision and
  Pattern Recognition}, 2020, pp. 9729--9738.

\bibitem{chen2020simple}
Ting Chen, Simon Kornblith, Mohammad Norouzi, and Geoffrey Hinton,
\newblock ``A simple framework for contrastive learning of visual
  representations,''
\newblock in {\em International conference on machine learning}. PMLR, 2020,
  pp. 1597--1607.

\bibitem{wang2021dense}
Xinlong Wang, Rufeng Zhang, Chunhua Shen, Tao Kong, and Lei Li,
\newblock ``Dense contrastive learning for self-supervised visual
  pre-training,''
\newblock in {\em Proceedings of the IEEE/CVF Conference on Computer Vision and
  Pattern Recognition}, 2021, pp. 3024--3033.

\bibitem{van2018representation}
Aaron Van~den Oord, Yazhe Li, and Oriol Vinyals,
\newblock ``Representation learning with contrastive predictive coding,''
\newblock {\em arXiv e-prints}, pp. arXiv--1807, 2018.

\bibitem{hadsell2006dimensionality}
Raia Hadsell, Sumit Chopra, and Yann LeCun,
\newblock ``Dimensionality reduction by learning an invariant mapping,''
\newblock in {\em IEEE Computer Society Conference on Computer Vision and
  Pattern Recognition (CVPR'06)}. IEEE, 2006, vol.~2, pp. 1735--1742.

\bibitem{caron2020unsupervised}
Mathilde Caron, Ishan Misra, Julien Mairal, Priya Goyal, Piotr Bojanowski, and
  Armand Joulin,
\newblock ``Unsupervised learning of visual features by contrasting cluster
  assignments,''
\newblock {\em Advances in Neural Information Processing Systems}, vol. 33, pp.
  9912--9924, 2020.

\bibitem{li2020prototypical}
Junnan Li, Pan Zhou, Caiming Xiong, and Steven Hoi,
\newblock ``Prototypical contrastive learning of unsupervised
  representations,''
\newblock in {\em International Conference on Learning Representations}, 2020.

\bibitem{li2021contrastive}
Yunfan Li, Peng Hu, Zitao Liu, Dezhong Peng, Joey~Tianyi Zhou, and Xi~Peng,
\newblock ``Contrastive clustering,''
\newblock in {\em Proceedings of the AAAI Conference on Artificial
  Intelligence}, 2021, vol.~35, pp. 8547--8555.

\bibitem{dosovitskiy2015discriminative}
Alexey Dosovitskiy, Philipp Fischer, Jost~Tobias Springenberg, Martin
  Riedmiller, and Thomas Brox,
\newblock ``Discriminative unsupervised feature learning with exemplar
  convolutional neural networks,''
\newblock {\em IEEE transactions on pattern analysis and machine intelligence},
  vol. 38, no. 9, pp. 1734--1747, 2015.

\bibitem{wu2018unsupervised}
Zhirong Wu, Yuanjun Xiong, Stella~X Yu, and Dahua Lin,
\newblock ``Unsupervised feature learning via non-parametric instance
  discrimination,''
\newblock in {\em Proceedings of the IEEE conference on computer vision and
  pattern recognition}, 2018, pp. 3733--3742.

\bibitem{ye2019unsupervised}
Mang Ye, Xu~Zhang, Pong~C Yuen, and Shih-Fu Chang,
\newblock ``Unsupervised embedding learning via invariant and spreading
  instance feature,''
\newblock in {\em Proceedings of the IEEE/CVF Conference on Computer Vision and
  Pattern Recognition}, 2019, pp. 6210--6219.

\bibitem{chen2020improved}
Xinlei Chen, Haoqi Fan, Ross Girshick, and Kaiming He,
\newblock ``Improved baselines with momentum contrastive learning,''
\newblock {\em arXiv preprint arXiv:2003.04297}, 2020.

\bibitem{sharma2020clustering}
Vivek Sharma, Makarand Tapaswi, M~Saquib Sarfraz, and Rainer Stiefelhagen,
\newblock ``Clustering based contrastive learning for improving face
  representations,''
\newblock in {\em 15th IEEE International Conference on Automatic Face and
  Gesture Recognition (FG 2020)}. IEEE, 2020, pp. 109--116.

\bibitem{liu2021simtriplet}
Quan Liu, Peter~C Louis, Yuzhe Lu, Aadarsh Jha, Mengyang Zhao, Ruining Deng,
  Tianyuan Yao, Joseph~T Roland, Haichun Yang, Shilin Zhao, et~al.,
\newblock ``Simtriplet: Simple triplet representation learning with a single
  gpu,''
\newblock in {\em International Conference on Medical Image Computing and
  Computer-Assisted Intervention}. Springer, 2021, pp. 102--112.

\bibitem{yang2021self}
Pengshuai Yang, Zhiwei Hong, Xiaoxu Yin, Chengzhan Zhu, and Rui Jiang,
\newblock ``Self-supervised visual representation learning for
  histopathological images,''
\newblock in {\em International Conference on Medical Image Computing and
  Computer-Assisted Intervention}. Springer, 2021, pp. 47--57.

\bibitem{zhao2021unsupervised}
Ziteng Zhao and Guanyu Yang,
\newblock ``Unsupervised contrastive learning of radiomics and deep features
  for label-efficient tumor classification,''
\newblock in {\em International Conference on Medical Image Computing and
  Computer-Assisted Intervention}. Springer, 2021, pp. 252--261.

\bibitem{wilson2020survey}
Garrett Wilson and Diane~J Cook,
\newblock ``A survey of unsupervised deep domain adaptation,''
\newblock {\em ACM Transactions on Intelligent Systems and Technology (TIST)},
  vol. 11, no. 5, pp. 1--46, 2020.

\bibitem{ahn2020unsupervised}
Euijoon Ahn, Ashnil Kumar, Michael Fulham, Dagan Feng, and Jinman Kim,
\newblock ``Unsupervised domain adaptation to classify medical images using
  zero-bias convolutional auto-encoders and context-based feature
  augmentation,''
\newblock {\em IEEE transactions on medical imaging}, vol. 39, no. 7, pp.
  2385--2394, 2020.

\bibitem{yang2020unsupervised}
Heran Yang, Jian Sun, Aaron Carass, Can Zhao, Junghoon Lee, Jerry~L Prince, and
  Zongben Xu,
\newblock ``Unsupervised mr-to-ct synthesis using structure-constrained
  cyclegan,''
\newblock {\em IEEE transactions on medical imaging}, vol. 39, no. 12, pp.
  4249--4261, 2020.

\bibitem{8494797}
Yuankai Huo, Zhoubing Xu, Hyeonsoo Moon, Shunxing Bao, Albert Assad, Tamara~K.
  Moyo, Michael~R. Savona, Richard~G. Abramson, and Bennett~A. Landman,
\newblock ``Synseg-net: Synthetic segmentation without target modality ground
  truth,''
\newblock {\em IEEE Transactions on Medical Imaging}, vol. 38, no. 4, pp.
  1016--1025, 2019.

\bibitem{zhang2018translating}
Zizhao Zhang, Lin Yang, and Yefeng Zheng,
\newblock ``Translating and segmenting multimodal medical volumes with
  cycle-and shape-consistency generative adversarial network,''
\newblock in {\em Proceedings of the IEEE conference on computer vision and
  pattern Recognition}, 2018, pp. 9242--9251.

\bibitem{zhou2021anatomy}
Bo~Zhou, Zachary Augenfeld, Julius Chapiro, S~Kevin Zhou, Chi Liu, and James~S
  Duncan,
\newblock ``Anatomy-guided multimodal registration by learning segmentation
  without ground truth: Application to intraprocedural cbct/mr liver
  segmentation and registration,''
\newblock {\em Medical image analysis}, vol. 71, pp. 102041, 2021.

\bibitem{chen2020anatomy}
Xu~Chen, Chunfeng Lian, Li~Wang, Hannah Deng, Tianshu Kuang, Steve Fung, Jaime
  Gateno, Pew-Thian Yap, James~J Xia, and Dinggang Shen,
\newblock ``Anatomy-regularized representation learning for cross-modality
  medical image segmentation,''
\newblock {\em IEEE Transactions on Medical Imaging}, vol. 40, no. 1, pp.
  274--285, 2021.

\bibitem{akagunduz2019defining}
Erdem Akagunduz, Adrian~G Bors, and Karla~K Evans,
\newblock ``Defining image memorability using the visual memory schema,''
\newblock {\em IEEE transactions on pattern analysis and machine intelligence},
  vol. 42, no. 9, pp. 2165--2178, 2019.

\bibitem{xu2019efficient}
Xuanang Xu, Fugen Zhou, Bo~Liu, Dongshan Fu, and Xiangzhi Bai,
\newblock ``Efficient multiple organ localization in ct image using 3d region
  proposal network,''
\newblock {\em IEEE transactions on medical imaging}, vol. 38, no. 8, pp.
  1885--1898, 2019.

\bibitem{zhou2021review}
S~Kevin Zhou, Hayit Greenspan, Christos Davatzikos, James~S Duncan, Bram
  Van~Ginneken, Anant Madabhushi, Jerry~L Prince, Daniel Rueckert, and Ronald~M
  Summers,
\newblock ``A review of deep learning in medical imaging: Imaging traits,
  technology trends, case studies with progress highlights, and future
  promises,''
\newblock {\em Proceedings of the IEEE}, vol. 109, no. 5, pp. 820--838, 2021.

\bibitem{lei2021contrastive}
Wenhui Lei, Wei Xu, Ran Gu, Hao Fu, Shaoting Zhang, Shichuan Zhang, and Guotai
  Wang,
\newblock ``Contrastive learning of relative position regression for one-shot
  object localization in 3d medical images,''
\newblock in {\em International Conference on Medical Image Computing and
  Computer-Assisted Intervention}. Springer, 2021, pp. 155--165.

\bibitem{yao2021one}
Qingsong Yao, Quan Quan, Li~Xiao, and S~Kevin~Zhou,
\newblock ``One-shot medical landmark detection,''
\newblock in {\em International Conference on Medical Image Computing and
  Computer-Assisted Intervention}. Springer, 2021, pp. 177--188.

\bibitem{hinton2015distilling}
Geoffrey Hinton, Oriol Vinyals, and Jeff Dean,
\newblock ``Distilling the knowledge in a neural network,''
\newblock {\em arXiv preprint arXiv:1503.02531}, 2015.

\bibitem{ronneberger2015u}
Olaf Ronneberger, Philipp Fischer, and Thomas Brox,
\newblock ``U-net: Convolutional networks for biomedical image segmentation,''
\newblock in {\em International Conference on Medical image computing and
  computer-assisted intervention}. Springer, 2015, pp. 234--241.

\bibitem{zhang2018mixup}
Hongyi Zhang, Moustapha Cisse, Yann~N Dauphin, and David Lopez-Paz,
\newblock ``mixup: Beyond empirical risk minimization,''
\newblock in {\em International Conference on Learning Representations}, 2018.

\bibitem{verma2019manifold}
Vikas Verma, Alex Lamb, Christopher Beckham, Amir Najafi, Ioannis Mitliagkas,
  David Lopez-Paz, and Yoshua Bengio,
\newblock ``Manifold mixup: Better representations by interpolating hidden
  states,''
\newblock in {\em International Conference on Machine Learning}. PMLR, 2019,
  pp. 6438--6447.

\bibitem{mackay2003information}
David~JC MacKay, David~JC Mac~Kay, et~al.,
\newblock {\em Information theory, inference and learning algorithms},
\newblock Cambridge university press, 2003.

\bibitem{goodfellow2014generative}
Ian Goodfellow, Jean Pouget-Abadie, Mehdi Mirza, Bing Xu, David Warde-Farley,
  Sherjil Ozair, Aaron Courville, and Yoshua Bengio,
\newblock ``Generative adversarial nets,''
\newblock {\em Advances in neural information processing systems}, vol. 27,
  2014.

\bibitem{kendall2017uncertainties}
Alex Kendall and Yarin Gal,
\newblock ``What uncertainties do we need in bayesian deep learning for
  computer vision?,''
\newblock {\em Advances in Neural Information Processing Systems}, vol. 30, pp.
  5574--5584, 2017.

\bibitem{bian2020uncertainty}
Cheng Bian, Chenglang Yuan, Jiexiang Wang, Meng Li, Xin Yang, Shuang Yu, Kai
  Ma, Jin Yuan, and Yefeng Zheng,
\newblock ``Uncertainty-aware domain alignment for anatomical structure
  segmentation,''
\newblock {\em Medical Image Analysis}, vol. 64, pp. 101732, 2020.

\bibitem{snell2017prototypical}
Jake Snell, Kevin Swersky, and Richard Zemel,
\newblock ``Prototypical networks for few-shot learning,''
\newblock in {\em Proceedings of the 31st International Conference on Neural
  Information Processing Systems}, 2017, pp. 4080--4090.

\bibitem{cuadros2009eyepacs}
Jorge Cuadros and George Bresnick,
\newblock ``Eyepacs: an adaptable telemedicine system for diabetic retinopathy
  screening,''
\newblock {\em Journal of diabetes science and technology}, vol. 3, no. 3, pp.
  509--516, 2009.

\bibitem{fraz2012ensemble}
Muhammad~Moazam Fraz, Paolo Remagnino, Andreas Hoppe, Bunyarit Uyyanonvara,
  Alicja~R Rudnicka, Christopher~G Owen, and Sarah~A Barman,
\newblock ``An ensemble classification-based approach applied to retinal blood
  vessel segmentation,''
\newblock {\em IEEE Transactions on Biomedical Engineering}, vol. 59, no. 9,
  pp. 2538--2548, 2012.

\bibitem{budai2013robust}
Attila Budai, R{\"u}diger Bock, Andreas Maier, Joachim Hornegger, and Georg
  Michelson,
\newblock ``Robust vessel segmentation in fundus images,''
\newblock {\em International journal of biomedical imaging}, vol. 2013, 2013.

\bibitem{hu2013automated}
Qiao Hu, Michael~D Abr{\`a}moff, and Mona~K Garvin,
\newblock ``Automated separation of binary overlapping trees in low-contrast
  color retinal images,''
\newblock in {\em International conference on medical image computing and
  computer-assisted intervention}. Springer, 2013, pp. 436--443.

\bibitem{hoover2000locating}
AD~Hoover, Valentina Kouznetsova, and Michael Goldbaum,
\newblock ``Locating blood vessels in retinal images by piecewise threshold
  probing of a matched filter response,''
\newblock {\em IEEE Transactions on Medical imaging}, vol. 19, no. 3, pp.
  203--210, 2000.

\bibitem{hoover2003locating}
Adam Hoover and Michael Goldbaum,
\newblock ``Locating the optic nerve in a retinal image using the fuzzy
  convergence of the blood vessels,''
\newblock {\em IEEE transactions on medical imaging}, vol. 22, no. 8, pp.
  951--958, 2003.

\bibitem{sivaswamy2015comprehensive}
Jayanthi Sivaswamy, S~Krishnadas, Arunava Chakravarty, G~Joshi, A~Syed Tabish,
  et~al.,
\newblock ``A comprehensive retinal image dataset for the assessment of
  glaucoma from the optic nerve head analysis,''
\newblock {\em JSM Biomedical Imaging Data Papers}, vol. 2, no. 1, pp. 1004,
  2015.

\bibitem{sivaswamy2014drishti}
Jayanthi Sivaswamy, SR~Krishnadas, Gopal~Datt Joshi, Madhulika Jain, and
  A~Ujjwaft~Syed Tabish,
\newblock ``Drishti-gs: Retinal image dataset for optic nerve head (onh)
  segmentation,''
\newblock in {\em 2014 IEEE 11th international symposium on biomedical imaging
  (ISBI)}. IEEE, 2014, pp. 53--56.

\bibitem{h25w98-18}
Prasanna Porwal, Samiksha Pachade, Ravi Kamble, Manesh Kokare, Girish Deshmukh,
  Vivek Sahasrabuddhe, and Fabrice Meriaudeau,
\newblock ``Indian diabetic retinopathy image dataset (idrid),'' 2018.

\bibitem{wang2017chestx}
Xiaosong Wang, Yifan Peng, Le~Lu, Zhiyong Lu, Mohammadhadi Bagheri, and
  Ronald~M Summers,
\newblock ``Chestx-ray8: Hospital-scale chest x-ray database and benchmarks on
  weakly-supervised classification and localization of common thorax
  diseases,''
\newblock in {\em Proceedings of the IEEE conference on computer vision and
  pattern recognition}, 2017, pp. 2097--2106.

\bibitem{van2006segmentation}
Bram Van~Ginneken, Mikkel~B Stegmann, and Marco Loog,
\newblock ``Segmentation of anatomical structures in chest radiographs using
  supervised methods: a comparative study on a public database,''
\newblock {\em Medical image analysis}, vol. 10, no. 1, pp. 19--40, 2006.

\bibitem{simonyan2014very}
Karen Simonyan and Andrew Zisserman,
\newblock ``Very deep convolutional networks for large-scale image
  recognition,''
\newblock {\em arXiv preprint arXiv:1409.1556}, 2014.

\bibitem{caron2018deep}
Mathilde Caron, Piotr Bojanowski, Armand Joulin, and Matthijs Douze,
\newblock ``Deep clustering for unsupervised learning of visual features,''
\newblock in {\em Proceedings of the European conference on computer vision
  (ECCV)}, 2018, pp. 132--149.

\bibitem{chen2021jigsaw}
Pengguang Chen, Shu Liu, and Jiaya Jia,
\newblock ``Jigsaw clustering for unsupervised visual representation
  learning,''
\newblock in {\em Proceedings of the IEEE/CVF Conference on Computer Vision and
  Pattern Recognition}, 2021, pp. 11526--11535.

\bibitem{biffi2018learning}
Carlo Biffi, Ozan Oktay, Giacomo Tarroni, Wenjia Bai, Antonio~De Marvao,
  Georgia Doumou, Martin Rajchl, Reem Bedair, Sanjay Prasad, Stuart Cook,
  et~al.,
\newblock ``Learning interpretable anatomical features through deep generative
  models: Application to cardiac remodeling,''
\newblock in {\em International conference on medical image computing and
  computer-assisted intervention}. Springer, 2018, pp. 464--471.

\bibitem{zhou2020comparing}
Hong-Yu Zhou, Shuang Yu, Cheng Bian, Yifan Hu, Kai Ma, and Yefeng Zheng,
\newblock ``Comparing to learn: Surpassing imagenet pretraining on radiographs
  by comparing image representations,''
\newblock in {\em International Conference on Medical Image Computing and
  Computer-Assisted Intervention}. Springer, 2020, pp. 398--407.

\bibitem{8454781}
Mingsheng Long, Yue Cao, Zhangjie Cao, Jianmin Wang, and Michael~I. Jordan,
\newblock ``Transferable representation learning with deep adaptation
  networks,''
\newblock {\em IEEE Transactions on Pattern Analysis and Machine Intelligence},
  vol. 41, no. 12, pp. 3071--3085, 2019.

\bibitem{he2016deep}
Kaiming He, Xiangyu Zhang, Shaoqing Ren, and Jian Sun,
\newblock ``Deep residual learning for image recognition,''
\newblock in {\em Proceedings of the IEEE conference on computer vision and
  pattern recognition}, 2016, pp. 770--778.

\bibitem{dong2019synthetic}
Xue Dong, Yang Lei, Sibo Tian, Tonghe Wang, Pretesh Patel, Walter~J Curran,
  Ashesh~B Jani, Tian Liu, and Xiaofeng Yang,
\newblock ``Synthetic mri-aided multi-organ segmentation on male pelvic ct
  using cycle consistent deep attention network,''
\newblock {\em Radiotherapy and Oncology}, vol. 141, pp. 192--199, 2019.

\bibitem{sandfort2019data}
Veit Sandfort, Ke~Yan, Perry~J Pickhardt, and Ronald~M Summers,
\newblock ``Data augmentation using generative adversarial networks (cyclegan)
  to improve generalizability in ct segmentation tasks,''
\newblock {\em Scientific reports}, vol. 9, no. 1, pp. 1--9, 2019.

\bibitem{bradski2008learning}
Gary Bradski and Adrian Kaehler,
\newblock {\em Learning OpenCV: Computer vision with the OpenCV library},
\newblock " O'Reilly Media, Inc.", 2008.

\end{thebibliography}
\end{document}